\documentclass[10pt,twocolumn,letterpaper]{article}

\usepackage{iccv}
\usepackage{times}
\usepackage{epsfig}
\usepackage{graphicx}
\usepackage{amsmath}
\usepackage{amssymb}
\usepackage{diagbox}
\usepackage{array}
\usepackage{enumitem}
\usepackage{multirow}
\usepackage{xcolor,colortbl}

\usepackage{upgreek}

\usepackage{pifont}%
\newcommand{\cmark}{\ding{51}}%
\newcommand{\xmark}{\ding{55}}%

\newcommand{\protein}[1]{\texttt{#1}}

\newcommand{\xdownarrow}[1]{%
    {\left\downarrow\vbox to #1{}\right.\kern-\nulldelimiterspace}
}
\renewcommand{\vec}[1]{\boldsymbol{#1}}
\newcommand{\E}{{\rm I\kern-.3em E}}
\renewcommand{\P}{{\rm I\kern-.3em P}}
\newcommand{\R}{\mathbb{R}}

\usepackage[pagebackref=true,breaklinks=true,letterpaper=true,colorlinks,bookmarks=false]{hyperref}

\iccvfinalcopy %

\def\iccvPaperID{606} %

\ificcvfinal\fi
\begin{document}

\title{GANs for Biological Image Synthesis}

\author{Anton Osokin\\
INRIA/ENS\thanks{DI \'{E}cole normale sup\'{e}rieure, CNRS, PSL Research University, Paris}, France\\
HSE\thanks{National Research University Higher School of Economics, Moscow}, Russia
\and
Anatole Chessel\\
\'{E}cole Polytechnique\thanks{LOB, \'{E}cole Polytechnique, CNRS, INSERM, Universit\'{e} Paris-Saclay},\\ France
\and
Rafael E. Carazo Salas\\
University of Bristol, UK
\and
Federico Vaggi\\
ENS\footnotemark[1], France\\
Amazon, USA
}

\maketitle
\begin{abstract}
In this paper, we propose a novel application of Generative Adversarial Networks (GAN) to the synthesis of cells imaged by fluorescence microscopy. Compared to natural images, cells tend to have a simpler and more geometric global structure that facilitates image generation. However, the correlation between the spatial pattern of different fluorescent proteins reflects important biological functions, and synthesized images have to capture these relationships to be relevant for biological applications. We adapt GANs to the task at hand and propose new models with casual dependencies between image channels that can generate multi-channel images, which would be impossible to obtain experimentally. We evaluate our approach using two independent techniques and compare it against sensible baselines. Finally, we demonstrate that by interpolating across the latent space we can mimic the known changes in protein localization that occur through time during the cell cycle, allowing us to predict temporal evolution from static images.
\end{abstract}

\section{Introduction}
In the life sciences, the last 20 years saw the rise of light fluorescence microscopy as a powerful way to probe biological events in living cells and organisms with unprecedented resolution.  The need to analyze quantitatively this deluge of data has given rise to the field of bioimage informatics \cite{Meijering2016} and is the source of numerous interesting and novel data analysis problems, which current machine learning developments could, in principle, help solve.

Generative models of natural images are among the most long-standing and challenging goals in computer vision.  Recently, the community has made significant progress in this task by adopting neural network machinery.  Examples of recent models include
denoising autoencoders~\cite{Bengio2013},
variational autoencoders~\cite{Kingma2014vae}, PixelCNNs~\cite{Oord2016conditional}
and Generative Adversarial Networks (GANs)~\cite{Goodfellow2014gan}.

\begin{figure}
    \vspace{-0.2cm}\includegraphics[trim = 2mm 32mm 7mm 5mm, clip, width=\columnwidth]{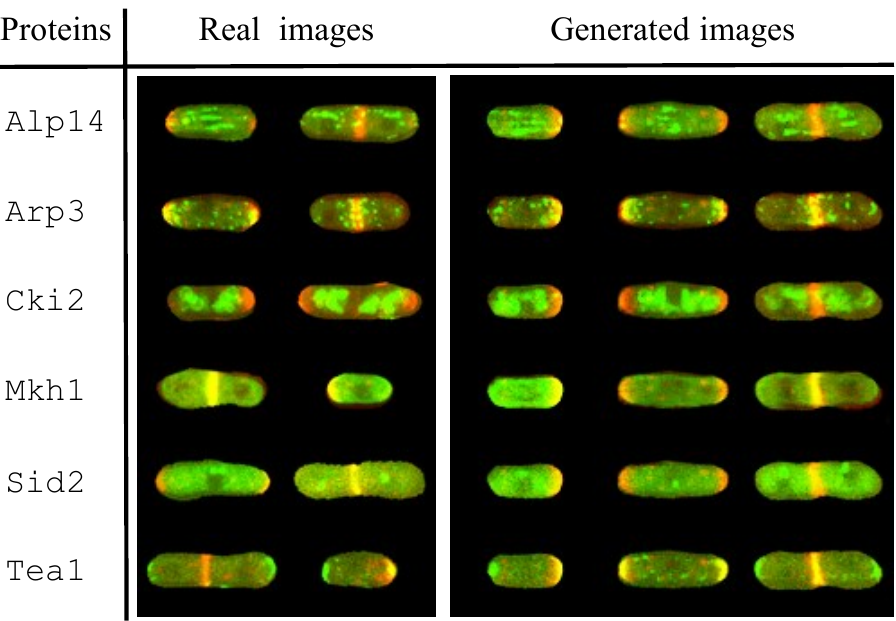}
    \caption{Real (left) and generated (right) images of fission yeast cells with protein \protein{Bgs4} depicted in  the red channel and 6 other proteins depicted in the green channel.
        The synthetic images were generated with our star-shaped GAN.
        The star-shaped model can generate multiple green channels aligned with the same red channel whereas the training images have only one green channel.\vspace{-0.4cm}
        \label{fig:proteins}}
\end{figure}

GANs~\cite{Goodfellow2014gan} are family of successful models, which have recently received widespread attention.
Unlike most other generative models, GANs do not rely on training objectives connected to the log likelihood.  Instead, GAN training can be seen as a minimax game between two models: the \emph{generator} aims to output images similar to the training set given random noise; while the \emph{discriminator} aims to distinguish the output of the generator from the training set.

Originally, GANs were applied to the MNIST dataset of handwritten digits~\cite{lecun1998mnist,Goodfellow2014gan}.
The consequent DCGAN model~\cite{Radford2016dcgan} was applied to the CelebA dataset~\cite{liu2015celeba} of human faces, the LSUN~\cite{Yu2016lsun,Radford2016dcgan} and ImageNet~\cite{deng2009imagenet} datasets of natural images.
We are not aware of any works applying GANs to biological images.

We work with a recently created bioimage dataset used to extract functional relationships between proteins, called the LIN dataset~\cite{Dodgson2017} comprising 170,000 fluorescence microscopy images of cells. In the LIN dataset, each image corresponds to a cell and is composed of signals from two independent fluorescence imaging channels (``red'' and ``green''), corresponding to the two different proteins tagged with red or green-emitting fluorophores, respectively.

In the LIN dataset, the red channel signal always corresponds to a protein named \protein{Bgs4}, which localizes to the areas of active growth of cells.  The green channel signal instead corresponds to any of 41 different ``polarity factors'', that is proteins that mark specific areas of the cells' cortex that help define a cell's geometry. Polarity factors include proteins like \protein{Alp14}, \protein{Arp3}, \protein{Cki2}, \protein{Mkh1}, \protein{Sid2} or \protein{Tea1}  (see Figure~\ref{fig:proteins} for image examples), each of which controls  the same biological process ``cellular polarity'' albeit each in a slightly different way. Each of the green-labeled polarity factors was imaged independently of the others. The biological aim of the LIN study is to investigate how those polarity factors (or proteins) interact with one another.

In this paper, we present a novel application of GANs to generate biological images.  Specifically, we want to tackle two concrete limitations of large scale fluorescent imaging screens: we want to use the common information contained in the red channel to learn how to generate a cell with several of the green-labeled proteins together. 
This would allow us to artificially predict how the localizations of those (independently imaged) proteins might co-exist in cells if they had been imaged together and circumvent the current technical limitations of being able to only image a limited number of signal channels at the same time. Second, taking advantage of the relationship between \protein{Bgs4} and the cell cycle stage, we want to study the dynamical changes in cellular localization that proteins undergo through time as cells grow and divide. 

To accomplish this, we make several contributions.
We modify the standard DCGAN~\cite{Radford2016dcgan} architecture by substituting the interdependence of the channels  with the causal dependence of the green on the red, allowing us  to  observe multiple modes of green signal for a single red setting.  Observing the mode collapse effect of GANs~\cite{Metz2017unrolled,Tolstikhin2017adagan} for our separable architecture, we incorporate the recent Wasserstein GAN (WGAN-GP) objective~\cite{Arjovsky2017wgan,Gulrajani2017wgangp}.  We propose two approaches to generate multi-channel images: regular WGAN-GP trained on multi-channel images, where extra channels for training are mined by nearest neighbor search in the training set, and a novel star-shaped generator trained directly on the two-channel images.  We carefully evaluate our models using two quantitative techniques: the neural network two-sample test (combining ideas from~\cite{Lopez-Paz2017tests} and~\cite{Gulrajani2017wgangp}) and by reconstructing samples in a held out test set with the optimization approach of~\cite{Metz2017unrolled}.
For reproducibility, we make the source code and data available online.\footnote{\url{https://github.com/aosokin/biogans}}

This paper is organized as follows. In Section~\ref{sec:relwork}, we discuss related works.
Section~\ref{sec:bio} reviews the relevant biological background for our application.
In Section~\ref{sec:gans}, we review GANs and present our modeling contributions.
We present the experimental evaluation in Section~\ref{sec:experiments} and conclude in Section~\ref{sec:conclusion}.

\section{Related Work}
\label{sec:relwork}
\vspace{-1mm}\paragraph{Generative Adversarial networks (GANs).}
Since the seminal paper by Goodfellow \textit{et al.}~\cite{Goodfellow2014gan} of 2014 (see also~\cite{Goodfellow2017tutorial} for a detailed review), GANs are becoming an increasingly popular model for learning to generate with the loss functions \emph{learned} jointly with the model itself.
Models with adversarial losses have been used in a wide range of applications, such as image generation~\cite{denton2015,Radford2016dcgan}, domain adaptation~\cite{Ganin2016adaptation}, text-to-image synthesis~\cite{reed2016generative}, synthesis of 3D shapes~\cite{Wu2016_3dgan} and texture~\cite{Li2016texture}, image-to-image translation~\cite{Isola2016pix2pix}, image super resolution~\cite{Ledig2017superres} and even generating radiation patterns in particle physics~\cite{Oliveira2017physics}.
However, these models suffer from issues such as mode collapse and oscillations during training, making them challenging to use in practice.  The community is currently tackling these problems from multiple angles.
Extensive effort has been placed on carefully optimizing the architecture of the network ~\cite{Radford2016dcgan,Salimans2016impgans} and developing best practices to optimize the training procedure\footnote{\url{https://github.com/soumith/ganhacks}}\!\!.
Another active area of research is improving the training objective function~\cite{Nowozin2016fgan,Chen2016infogan,Poole2016,Zhao2017ebgan,Arjovsky2017wgan,Metz2017unrolled,Tolstikhin2017adagan,Gulrajani2017wgangp}.

In this paper, we build on the DCGAN architecture~\cite{Radford2016dcgan} combined with the Wasserstein loss~\cite{Arjovsky2017wgan,Gulrajani2017wgangp}, where the latter is used to help with the mode collapse issue, appearing especially in our separable setting.

\vspace{-3.5mm}\paragraph{Conditioning for GANs.}
Starting from conditioning on the class labels~\cite{Mirza2014cgan,denton2015,Odena2016,Dumoulin2017}, researchers have extended conditioning to user scribbles~\cite{Zhu2016manipulation} and images~\cite{Wang2016,Isola2016pix2pix,Zhu2017cycleGan}.
While the quality of images generated by~\cite{Wang2016,Isola2016pix2pix,Zhu2017cycleGan} is high, their models suffer from conditional mode collapse, i.e., given the first (source) image there is very little or no variety in the second (target image).
This effect might be related to the fact that the dataset contained only one target image available for each source image, so the model has only indirect supervision for generating multiple conditioned images.
We have applied the pix2pix method of~\cite{Isola2016pix2pix} to the LIN dataset and it learned to produce high-quality green images given the red input. However, it was unable to generate multiple realistic green images for one red input.

Given the difficulty in learning robust latent spaces when conditioning on an image, we opted for an alternate approach.  We propose a new architecture for the generator, where the red channel and green channels are given independent random noise, and only the red channel is allowed to influence the green channel, see Figure~\ref{fig:arch} (right).

\vspace{-3.5mm}\paragraph{Factors of variation.} Chen \textit{et al.}~\cite{Chen2016infogan} and Mathieu \textit{et al.}~\cite{Mathieu2016disentangling} used unsupervised methods that encourage disentangling factors of variation in the learned latent spaces, e.g., separating the numerical value of a handwritten digit from its writing style.
In contrast to these works, we do not rely on unsupervised training to discover factors of variations, but explicitly embed the separation into the model.

\vspace{-2mm}\paragraph{Analysis and synthesis of biological images.}
With large scale imaging studies becoming more common in biology, the automated analysis of images is now crucial in many studies to prove the existence of an effect, process large datasets or link with models and simulation~\cite{Meijering2016, chessel2017overview}.  Although the field has only recently embraced deep learning, neural networks are now starting to make a splash, mainly in classical discriminative settings~\cite{VanValen2016}.

While, to our knowledge, this work is the first reported use of GANs on samples from fluorescent microscopy, generative models have been widely used in biology \cite{Murphy2015}.
For example, Johnson \textit{et al}~\cite{Johnson2015a} learned to generate punctuate patterns in cells (conditional on microtubule localization) showing the potential of those methods in studying the relative sub-cellular positions of several proteins of interest.

Recently, sharing of large biological datasets has greatly improved~\cite{ljosa2012annotated}.  Further, EBI has made a large investment to develop the IDR (Image Data Resource)~\cite{Williams089359}, a database built on top of open source tools to facilitate the sharing of terabyte sized datasets with complex metadata.

\section{Biological Background}
\label{sec:bio}
\subsection{Fluorescent Imaging}
Fluorescence microscopy is based on fluorescent compounds, i.e., compounds which can absorb light at given wavelength (the \emph{absorption spectrum}) and re-emit it almost immediately at a slightly different wavelength (the \emph{emission spectrum}). In the case of fluorescent proteins (FPs), of which the Green Fluorescent Protein (GFP)~\cite{chalfie1994gfp,tsien1998gfp} is the first and most widely used one, the fluorescing compound is attached to the protein of interest via genetic engineering. Many FPs of various absorption and emission spectra exist, e.g., Red Fluorescent Protein (RFP)~\cite{shaner2004improved}. By genetically tagging different proteins of interest with FPs of different color, one can image them in the same cell at the same time and thus investigate their co-localization.
However, the number of proteins that can be tagged and imaged at the same time is limited to 3-4 due to the limited number of FPs with non-overlapping absorption spectra.

Multi-channel fluorescent images are very different from natural images.
In natural images, 
color is determined by the illumination and the properties of a particular material in the scene.  In order
to generate realistic natural samples, a GAN must capture the relationship between the materials that make up a 
particular object and its hues.
In contrast, in fluorescent images, the intensity of light in a given channel corresponds to the local concentration of the tagged protein, and the correlation between signals in different channels represents important information about the relationship between proteins, but the color does not reflect any intrinsic property about the protein itself.

\subsection{Fission Yeast Cells}
\label{sec:biobackground}
Fission yeast (\emph{Schizosaccharomyces pombe}) cells are rod shaped unicellular eukaryotes with spherical hemisphere caps.  They are
born 7 $\upmu$m long and 4 $\upmu$m wide, and grow in length to 14 $\upmu$m while maintaining their width constant.  Newly born fission yeast cells
start by growing only at the pre-existing  end until they reach a critical size, and then switch to bipolar (from the two sides) growth. %
Bipolar growth continues until cells reach their final length, when they stop growing and start to form a
cytokinetic ring in the middle, which is responsible for cleaving the mother cells into two daughters~\cite{pollard2010understanding}.
Interestingly, for most of the cell cycle the length of the cell is a good proxy for its ``age'', i.e. the time it has spent growing since its ``birth''.

\protein{Bgs4}, the protein tagged in the red channel, is responsible for cell wall remodeling, and localizes to areas of active growth (see Figure~\ref{fig:proteins} for examples of images).
Thus, by observing \protein{Bgs4}, one can accurately
infer growth cycle stage, and predict where cell growth is occurring.

\subsection{The LIN Dataset}
All experiments in this paper make use of a recent dataset of images of fission yeast cells, which was originally produced to study polarity networks~\cite{Dodgson2017}.
The LIN dataset consists of around 170,000 of images, with each image being centered on one cell; cell segmentation was performed separately (see~\cite{Dodgson2017} for details) and the corresponding outline is also available. Each image is a 3D stack of 2D images where each pixel correspond to a physical size of 100nm; each $z$-plane is distant by 300nm. Every image is composed of two channels, informally called the ``red'' and the ``green'', where light emitted at a precise wavelength is recorded. In this dataset two types of fluorescent-tagged proteins are used: \protein{Bgs4} in the red channel, and one of 41 different polarity regulating proteins in the green channel. 
A full description of all tagged proteins is beyond the scope of this paper: we refer interested readers to~\cite{martin2014cell,Dodgson2017}.

In this paper, we concentrate on a subset of 6 different polarity factors, spanning a large set of different cellular localizations. This gives us 26,909 images of cells, which we, for simplicity, center crop and resize to resolution of $48\times 80$.

\section{GANs for Image Generation}
\label{sec:gans}
\subsection{Preliminaries}
\vspace{-1mm}\paragraph{GAN.} The framework of generative adversarial networks~\cite{Goodfellow2014gan,Goodfellow2017tutorial} is formulated as a minimax two-player game between two neural networks: \emph{generator} and \emph{discriminator}.
The generator constructs images given random noise whereas the discriminator tries to classify if its input image is real (from the training set) or fake (from the generator).
The goal of the generator is to trick the discriminator, such that it cannot easily classify.
The discriminator is often referred to as the adversarial loss for training the generator.

More formally, consider a data-generating distribution~$\P_{d}$ and a training set of images $\vec{x} \in \mathcal{X}$ coming from it.
The generator $G(\vec{z}; \theta_G)$ is a neural network parameterized by~$\theta_G$ that takes random noise $\vec{z}$ from distribution~$\P_{\vec{z}}$ as input and produces an image~$\vec{x}_{\text{fake}} \in \mathcal{X}$.
The discriminator~$D(\vec{x}; \theta_D)$ is a neural network parameterized by~$\theta_D$ that takes either a training image~$\vec{x}$ or a generated image~$\vec{x}_{\text{fake}}$ and outputs a number in the segment~$[0, 1]$, where zero is associated with fake images and one~-- with the real images.
As introduced in~\cite{Goodfellow2014gan}, the key quantity is the negative cross-entropy loss on the discriminator output:\vspace{-2mm}
\begin{multline}
\label{eq:ganLoss}
L(\theta_D, \theta_G) =  \E_{\vec{x}\sim \P_{\text{data}}} \log D(\vec{x}; \theta_D)
\\ + \E_{\vec{z}\sim \P_{\vec{z}}} \log( 1-D(G(\vec{z}; \theta_G); \theta_D) ).
\end{multline}
\vspace{-5mm}

\noindent The discriminator maximizes~\eqref{eq:ganLoss} w.r.t.~$\theta_D$ and the generator, at the same time, minimizes~\eqref{eq:ganLoss} w.r.t.~$\theta_G$.
In practice, both optimization tasks are attacked simultaneously by alternating between the steps of the two optimizers.

As noted by~\cite{Goodfellow2014gan}, the objective~$\log( 1-D(G(\vec{z}; \theta_G); \theta_D) )$ often leads to saturated gradients at the initial stages of the training process when the generator is ineffective, i.e., its samples are easy to discriminate from the real data.
One practical trick to avoid saturated gradients is to train the generator by maximizing~$\log D(G(\vec{z}; \theta_G); \theta_D)$ instead.

Goodfellow \emph{et al.}~\cite{Goodfellow2014gan} showed that the minimax formulation~\eqref{eq:ganLoss} can be reformulated via minimization of the Jensen-Shannon (JS) divergence\footnote{The Jensen-Shannon divergence between the two distributions is a symmetrized version of the Kullback-Leibler divergence, i.e., $JS(\P_{d} , \P_G) = \frac12 KL(\P_{d}\|\P_M) + \frac12 KL(\P_G\|\P_M)$, where $\P_M = \frac12 (\P_G +\P_{d})$ is the averaged distribution.} between the data-generating distribution~$\P_{d}$ and the distribution~$\P_G$ induced by $\P_{\vec{z}}$ and $G$.

\begin{figure}
    \centering
    \includegraphics[trim = 0mm 2.8mm 8mm 2.2mm, clip, width=\columnwidth]{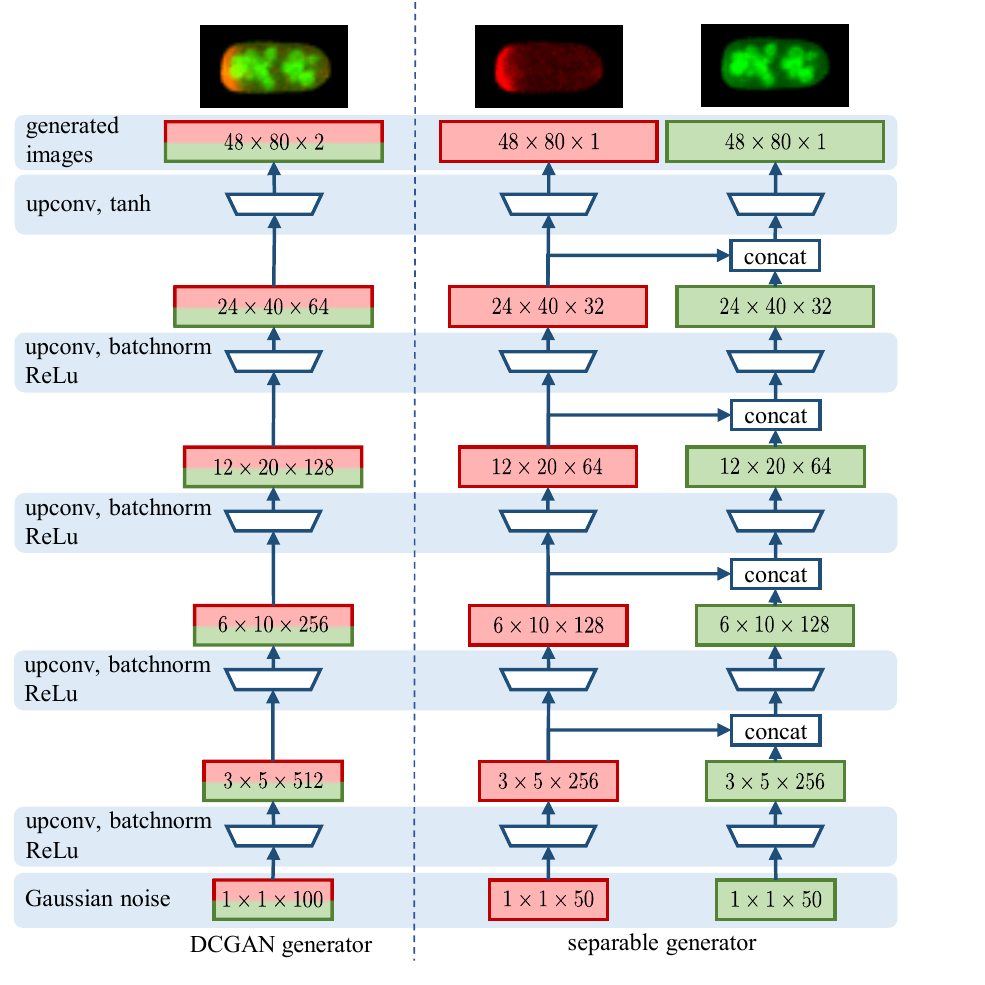}
    \caption{Architectures of the DCGAN generator (left) and our separable generator (right).\vspace{-0.1cm}\label{fig:arch}}
\end{figure}

For the architectures of both the generator and the discriminator, we largely reuse a successful version of Radford \emph{et al.}~\cite{Radford2016dcgan} called DCGAN.
The generator of DCGAN (see Figure~\ref{fig:arch}, left) is based on up-convolutions~\cite{Dosovitskiy2017chairs} interleaved with ReLu non-linearity and batch-normalization~\cite{Ioffe2015batchnorm}.
We refer to~\cite{Radford2016dcgan} for additional details.

\vspace{-3mm}\paragraph{Wasserstein GAN.}
Recently, Arjovsky \emph{et al.}~\cite{Arjovsky2017wgan} have demonstrated that in some cases the JS divergence behaves badly and cannot provide any useful direction for training, e.g., when it is discontinuous.
To overcome these degeneracies, they consider the earth mover's distance (equivalent to the 1-st Wasserstein distance) between the distributions\vspace{-2mm}
\begin{equation}
\label{eq:wasserOriginal}
W(\P_{d} , \P_G) = \inf_{ \P \in \Pi(\P_{d} , \P_G)} \E_{(\vec{x}, \vec{x}') \sim \P} \|\vec{x} - \vec{x}'\|,\vspace{-2mm}
\end{equation}
where set~$\Pi(\P_{d} , \P_G)$ is a set of all joint distributions~$\P$ on~$\vec{x}$ and $\vec{x}'$ whose marginals are~$\P_{d}$ and $\P_G$, respectively.
Intuitively, the distance~\eqref{eq:wasserOriginal} indicates the cost of the optimal movement of the probability mass from~$\P_{d}$ to~$\P_G$.
According to~\cite{Arjovsky2017wgan} by using duality, one can rewrite~\eqref{eq:wasserOriginal} as\vspace{-3mm}
\begin{equation}
\label{eq:wasserDual}
W(\P_{d} , \P_G) \!=\! \sup_{ D \in \mathcal{C}^1 } \Bigl( \E_{\vec{x} \sim \P_{d}}  D(\vec{x}) \!-\! \E_{\vec{x}' \sim \P_{G}}  D(\vec{x}')\Bigr),\vspace{-2mm}
\end{equation}
where~$\mathcal{C}^1$ is the set of all $1$-Lipschitz functions $D: \mathcal{X} \to \R$.

Optimizing w.r.t.\ the set~$\mathcal{C}^1$ is complicated.
As a practical approximation to the set of all $1$-Lipschitz functions, Arjovsky \emph{et al.}~\cite{Arjovsky2017wgan} suggest to use neural networks~$D(\vec{x}; \theta_D)$ with all parameters~$\theta_D$ clipped to a fixed segment. %
Very recently, Gulrajani \textit{et~al.}~\cite{Gulrajani2017wgangp} proposed a surrogate objective to~\eqref{eq:wasserDual}, which is based on the $L_2$-distance between the norm of the discriminator gradient at specific points and one.
In all, we arrive at the minimax game
\vspace{-3mm}
\begin{multline}
\label{eq:wassesPractical}
W(\theta_D, \theta_G) =  \E_{\vec{z}\sim \P_{\vec{z}}} D(G(\vec{z}; \theta_G); \theta_D) \\
- \E_{\vec{x}\sim \P_{\text{data}}} D(\vec{x}; \theta_D)  + R(\theta_D),
\end{multline}
\vspace{-6mm}

\noindent
where~$R$ is the regularizer (see~\cite{Gulrajani2017wgangp} for details).
The objective~\eqref{eq:wassesPractical} is very similar to the original game of GANs~\eqref{eq:ganLoss}, but has better convergence properties.
In what follows, we refer to the method of~\cite{Gulrajani2017wgangp} as WGAN-GP.

\subsection{Model Extensions}
\label{sec:multi-channel}
In this section, we present our modeling contributions.
First, we describe our approach to separate the red and green channels of the generator.
Second, we discuss a way to train a multi-channel generator using the two-channel data in the LIN dataset.   Finally, we propose a new star-shaped architecture that uses the red-green channel separation to obtain multiple channels in the output.

\vspace{-2mm}\paragraph{Channel separation.}
The key idea of the channel separation consists in separating the filters of all the up-convolutional layers and the corresponding features into two halves.  The first set of filters is responsible for generating the red channel, while the second half generates the green channel.  To make sure the green channel matches the red one, we use one way connections from the red convolutional filters towards the green ones. 
Figure~\ref{fig:arch} (right) depicts our modification in comparison to DCGAN (left).

\vspace{-3mm}\paragraph{Multi-channel models.}
The LIN dataset~\cite{Dodgson2017} contains only two-channel images, the red and one type of the green at a time.
Obtaining up to 4 channels simultaneously from a set of 40 proteins (a fixed red and 3 greens) would require the creation of nearly 60,000 yeast strains.  Scaling even higher is currently impossible with this imaging technique due to the limited number of FPs with non-overlapping absorption spectra.
Because of these constraints, training the generator only on a subset of channels is a task of practical importance.  The first approach we present consists in training a multi-channel GAN using an artificial training set of multi-channel images created from the real two-channel images.
We proceed as follows: for each two-channel image, we search in every other class for its nearest-neighbors (using $L_2$-distance) in the red channel. Then, we create a new sample by combining the original image with the green channels of its nearest neighbors in other classes.

We can then use this dataset to train a multi-output DCGAN.  The only difference in the architecture is that the generator outputs $c+1$ channels, where $c$ is the number of green channels used in the experiment, and the discriminator takes $(c+1)$-channel images as input.

\vspace{-3mm}\paragraph{Star-shaped model.}
In our experiments, the multi-channel approach did not perform well, because, even using the nearest neighbors, the extra greens channels were not exactly consistent with the original red signal, emphasizing the importance of correlations between channels.

To overcome this effect, we propose a star-shaped architecture for the generator, consisting of a single red tower (a stack of upconvolutional layers with non-linearities in-between) that feeds into $c$ green towers (see Figure~\ref{fig:arch}, right).   Unlike the multi-channel model described above, the green outputs are independent conditioned on the red. Thus, the model can be trained using the existing two-channel images.

In our experiments, we found it important to use batch normalization~\cite{ioffe2015batch} in the red tower only once, compared to a more naive way of $c$ times.
The latter leads to interference between several normalizations of the same features and prevents convergence of the training scheme.

After the forward pass, we use $c$ discriminators attached to different versions of the greens, all paired with the same generated red.
For the WGAN-GP version of this model, we apply the original procedure of~\cite{Gulrajani2017wgangp} with the modification that during the discriminator update we simultaneously update all~$c$ discriminators, and the generator receives back the accumulated gradient.

\section{Experiments}
\label{sec:experiments}
Evaluating generative models is in general non-trivial.  In the context of GANs and other likelihood-free approaches, evaluation is even harder, because the models do not provide a way to compute the log-likelihood on the test set, which is the most common evaluation technique.
Recently, a number of techniques applicable to evaluating GANs have been proposed~\cite{Lopez-Paz2017tests,Metz2017unrolled,Wu2017evaluation}.
Among those, we chose the following two: the neural-network two-sampled test discussed by~\cite{Lopez-Paz2017tests} combined with the surrogates of the earth mover's distance~\cite{Arjovsky2017wgan,Gulrajani2017wgangp} and an optimization-based approach of~\cite{Metz2017unrolled} to check if the test samples can be well reconstructed.
We modify these techniques to match our needs and check their performance using sensible baselines (Sections~\ref{sec:twoSampleTest} and~\ref{sec:exp:optimization}).
Finally, in Section~\ref{sec:cellCycle}, we show the cell growth cycle generated with our star-shaped model.

\subsection{Neural-network Two-sample Test}
\label{sec:twoSampleTest}
Lopez-Paz and Oquab~\cite{Lopez-Paz2017tests} have recently applied the classifier two-sample test (C2ST) to evaluate the quality of GAN models.
A trained generator is evaluated on a held-out test set.
This test test is split again into a test-train and test-test subsets.
The test-train set is then used to train a fresh discriminator, which tries to distinguish fake images (from the generator) from the real images.
Afterwards, the final measure of the quality of the generator is computed as the performance of the new discriminator on the test-test set and the freshly generated images.

When C2ST is applied for images, the discriminator is usually a ConvNet, but even very small ConvNets can discriminate between fake and real images almost perfectly.  To obtain a useful measure, Lopez-Paz and Oquab \cite{Lopez-Paz2017tests} deliberately weaken the ConvNet by fixing some of its parameters to the values obtained by pre-training on ImageNet.

ImageNet-based features are clearly not suitable for LIN cell images, so we weaken the discriminator in another way.
We use the negation of the WGAN-GP~\cite{Gulrajani2017wgangp} discriminator objective as a surrogate to the earth mover's distance.
Similar to~\cite{Lopez-Paz2017tests}, we train this discriminator on the test-train subset and compute the final estimates on the test-test subset.
For all the runs, we repeat the experiment on 10 different random splits of the test set and train the discriminator for 5000 steps with the optimizer used by~\cite{Gulrajani2017wgangp}.
For the experiments involving multi-channel generators, we train a separate discriminator for each green channel paired with the red channel.

\begin{figure}[t]
    \begin{center}
        \begin{tabular}{@{}m{2.5mm}@{\:}|@{\:}m{8mm}@{\:}|@{\:}m{8mm}@{\:}|@{\:}m{57.5mm}@{}}
            & Steps & C2ST & \parbox{55mm}{\centering Samples}
            \\[0mm]
            \hline
            \multirow{3}{*}{\rotatebox[origin=c]{90}{\parbox{23mm}{\centering sep.~GAN}}} & 1k & \parbox{8mm}{\small $11.0\pm$ \\[-0.5mm] $0.1$} &
            \vspace{1mm}\multirow{3}{*}{\includegraphics[width=57.5mm]{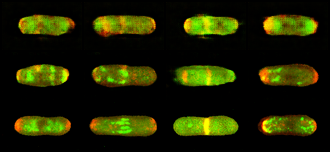}}
            \\[5mm]
            & 5k & \parbox{8mm}{\small $6.7\pm$ \\[-0.5mm] $0.1$}
            &
            \\[5mm]
            & 50k & \parbox{8mm}{\small $3.2\pm$ \\[-0.5mm] $0.1$} &
            \\[5mm]
            \hline
            \multirow{3}{*}{\rotatebox[origin=c]{90}{\parbox{23mm}{\centering sep.~WGAN-GP}}} & 1k & \parbox{8mm}{\small $6.0\pm$ \\[-0.5mm] $0.1$} &
            \vspace{1mm}\multirow{3}{*}{\includegraphics[width=57.5mm]{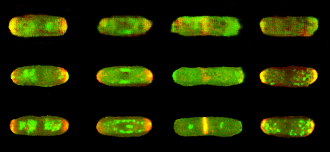}}
            \\[5mm]
            & 5k & \parbox{8mm}{\small $2.2\pm$ \\[-0.5mm] $0.1$} &
            \\[5mm]
            & 50k & \parbox{8mm}{\small $\mathbf{1.6\pm}$\\[-0.5mm]$\mathbf{0.1}$} &
            \\[5mm]
            \hline
            \rotatebox[origin=c]{90}{Real} & - & \parbox{8mm}{\small $-0.7$ \\[-0.5mm] $\pm0.6$} &
            \vspace{1mm}\includegraphics[width=57.5mm]{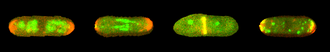}\\[-0.1cm]
        \end{tabular}
    \end{center}
    \vspace{-2mm}\caption{Scores of the classifier two-sample test (C2ST) between the generators and the hold-out test sets of images. We report the scores of separable GAN and WGAN-GP at different stages of training. For each line, we show the samples from the corresponding models to demonstrate that the lower C2ST scores correspond to better-looking (sharper, less artifacts, etc.) images. Best viewed in color and on a screen. An extended version of this figure is given in Appendix~\ref{app:c2st}.\vspace{-5mm}
        \label{fig:wdistEvolution}}
\end{figure}

\begin{figure}[t]
    \begin{tabular}{@{}m{4mm}@{\;}m{78mm}@{}}
        &  \parbox{78mm}{\centering Samples from separable models}\\
        \hline
        \rotatebox[origin=c]{90}{separable GAN}  &
        \vspace{1mm}\includegraphics[trim = 0mm 0mm 86mm 54mm, clip, width=78mm]{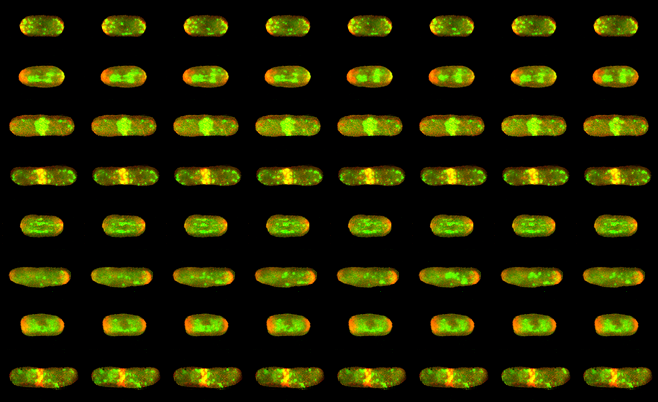}
        \\[0mm]
        \hline
        \rotatebox[origin=c]{90}{separable WGAN-GP}  &
        \vspace{1mm}\includegraphics[trim = 0mm 0mm 86mm 54mm, clip, width=78mm]{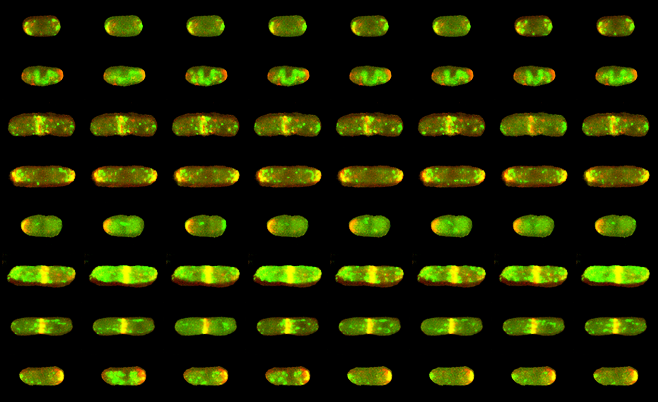}\\[-2mm]
    \end{tabular}
    \vspace{1mm}\caption{Samples generated by separable GAN (top) and WGAN-GP (bottom) models trained on the 6 selected proteins shown in Figure~\ref{fig:proteins}. Each row has samples with identical red channel, but different green ones. We observe that WGAN-GP provides much larger variability of the green channel conditioned on the red. In particular, in the three bottom rows, even the type of the protein changes, which we have never observed for the samples of GAN (this effect should be present, because the model is trained without any distinction between the classes, but is surprisingly rare). This difference is captured by the C2ST evaluation: the GAN model has a score of~$3.2\pm0.1$ compared to $1.6\pm 0.1$ of WGAN-GP.\vspace{-5mm}
        \label{fig:modeCollapse}}
\end{figure}

In our experiments, the training procedure occasionally failed and produced large outliers.  To be more robust, we always report a median over 10 random splits together with the median absolute deviation to represent the variance.
In Appendix \ref{app:c2st}, we additionally quantitatively and qualitatively compare the WGAN-GP~\cite{Gulrajani2017wgangp}, WGAN~\cite{Arjovsky2017wgan} and cross-entropy discriminators used in C2ST.

\vspace{-3mm}\paragraph{Sanity checks of the two-sample test.}
We evaluate C2ST in two baseline settings.  First, we compare the separable GAN~\cite{Radford2016dcgan} and the WGAN-GP~\cite{Gulrajani2017wgangp} models (based on the same DCGAN architecture,  trained on the same set of images of 6 proteins) at different stages of the training process.
For each of these models, we also show qualitative difference between the generated images. 
Figure~\ref{fig:wdistEvolution} shows that along the training process, quality of both GAN and WGAN-GP improves, i.e., generated images become sharper and contain less artifacts, consistent with the C2ST score.
To better visualize the difference between the trained GAN and WGAN-GP models, in Figure~\ref{fig:modeCollapse}, we show multiple samples of the green channel corresponding to the same red channel.
We see that the C2ST evaluation captures several aspects of the visual quality (such as sharpness, correct shape, absence of artifacts, diversity of samples) and provides a meaningful score.

From Figures~\ref{fig:wdistEvolution} and~\ref{fig:modeCollapse}, we also conclude that the quality of GAN samples is worse than the quality of WGAN-GP according to visual inspection. C2ST (based on WGAN-GP) confirms this observation, which is not surprising given that WGAN-GP was trained using the same methodology. Surprisingly, when evaluated with the cross-entropy C2ST, WGAN-GP also performs better than GAN (see Appendix~\ref{app:c2st} for details).

As the second baseline evaluation, we use C2ST to compare real images of different classes.  Table~\ref{tbl:wdist:realVsReal} shows that when evaluated w.r.t.\ the test set of the same class the estimates are significantly smaller (but with non-zero variance) compared to when evaluated w.r.t.\ different classes.
Note that the C2ST score is not a metric. In particular, Table~\ref{tbl:wdist:realVsReal} is not symmetric reflecting biases between the train/test splits. Specifically to WGAN-GP, the score can also be negative, because the quadratic regularization term is the dominant part of the objective~\eqref{eq:wassesPractical} when the two image sources are very similar.

As an additional test, we include two extra proteins ~\protein{Fim1} and~\protein{Tea4} that are known to have similar localization to ~\protein{Arp3} and~\protein{Tea1}, respectively.  We observe that C2ST reflects this similarity by giving the pairs of similar proteins a much smaller score compared to most of other pairs (but still significantly higher than comparing a protein to itself).

\begin{table*}
    \begin{center}
        \begin{tabular}{c|cccccc|cc}
            test & \protein{Alp14}  & \protein{Arp3}  & \protein{Cki2}  & \protein{Mkh1}  & \protein{Sid2}  & \protein{Tea1}  & \protein{Fim1}  & \protein{Tea4} \\
            \hline
            \protein{Alp14} & \cellcolor{gray!25}0.1 $\pm$ 0.2 & 12.5 $\pm$ 0.3 & 8.1 $\pm$ 0.3 & 12.5 $\pm$ 0.5 & 9.5 $\pm$ 0.2 & 10.9 $\pm$ 0.3 & 15.6 $\pm$ 0.3 & 11.4 $\pm$ 0.3 \\
            \protein{Arp3} & 14.4 $\pm$ 0.2 & \cellcolor{gray!25}0.8 $\pm$ 0.4 & 16.2 $\pm$ 0.2 & 11.5 $\pm$ 0.4 & 20.5 $\pm$ 0.3 & 13.2 $\pm$ 0.2 & \textbf{3.7 $\pm$ 0.2} & 18.3 $\pm$ 0.3 \\
            \protein{Cki2} & 8.6 $\pm$ 0.2 & 15.9 $\pm$ 0.3 & \cellcolor{gray!25}-0.2 $\pm$ 0.3 & 13.7 $\pm$ 0.4 & 12.0 $\pm$ 0.3 & 15.8 $\pm$ 0.3 & 18.5 $\pm$ 0.4 & 16.0 $\pm$ 0.5 \\
            \protein{Mkh1} & 12.3 $\pm$ 0.4 & 12.2 $\pm$ 0.6 & 13.6 $\pm$ 0.3 & \cellcolor{gray!25}-0.2 $\pm$ 0.4 & 12.4 $\pm$ 0.6 & 13.3 $\pm$ 0.6 & 15.1 $\pm$ 0.5 & 14.9 $\pm$ 0.8 \\
            \protein{Sid2} & 9.0 $\pm$ 0.3 & 19.5 $\pm$ 0.4 & 11.8 $\pm$ 0.5 & 13.4 $\pm$ 0.9 & \cellcolor{gray!25}-0.6 $\pm$ 0.3 & 12.6 $\pm$ 0.3 & 23.9 $\pm$ 0.4 & 7.7 $\pm$ 0.6 \\
            \protein{Tea1} & 11.3 $\pm$ 0.3 & 11.5 $\pm$ 0.5 & 15.9 $\pm$ 0.3 & 14.4 $\pm$ 0.6 & 13.1 $\pm$ 0.1 & \cellcolor{gray!25}-0.1 $\pm$ 0.4 & 14.5 $\pm$ 0.5 & \textbf{6.9 $\pm$ 0.5} \\
            \hline
            \protein{Fim1} & 16.3 $\pm$ 0.2 & \textbf{2.8 $\pm$ 0.3} & 18.4 $\pm$ 0.2 & 14.5 $\pm$ 0.3 & 23.4 $\pm$ 0.3 & 15.1 $\pm$ 0.2 & \cellcolor{gray!25}-0.2 $\pm$ 0.3 & 20.8 $\pm$ 0.5 \\
            \protein{Tea4} & 9.7 $\pm$ 0.6 & 15.8 $\pm$ 0.7 & 14.0 $\pm$ 0.9 & 13.9 $\pm$ 0.9 & 6.2 $\pm$ 0.4 & \textbf{5.9 $\pm$ 0.3} & 19.5 $\pm$ 0.7 & \cellcolor{gray!25}-0.5 $\pm$ 0.7 \\[-0.2cm]
        \end{tabular}
    \end{center}
    \caption{ Results of C2ST with WGAN-GP when comparing real images of different proteins.
        For each run, the training images of one class are evaluated w.r.t.\ the test images of another class.
        The reported values are comparable with Table~\ref{tbl:wdist:wgansVsReal}, but not with Figure~\ref{fig:wdistEvolution}.\vspace{-0.2cm}
        \label{tbl:wdist:realVsReal}}
\end{table*}

\begin{table*}
    \begin{center}
        \begin{tabular}{@{}p{28mm}|p{16.5mm}|p{22mm}|p{20mm}p{20mm}p{20mm}p{20mm}@{}}
            & real images & \parbox{20mm}{one-class\\ non-separable} & \parbox{15mm}{one-class\\ separable} & \parbox{20mm}{multi-channel non-separable} & \parbox{20mm}{multi-channel separable} & star-shaped \\
            \hline
            separable red/green & - & \xmark & \cmark & \xmark & \cmark & \cmark  \\
            class conditioned   & - & \xmark & \xmark & \cmark & \cmark & \cmark  \\
            \hline
            \protein{Alp14} &  0.1 $\pm$ 0.2 &  \textbf{0.6 $\pm$ 0.3} & 1.2 $\pm$ 0.2 & 3.2 $\pm$ 0.4 & 2.3 $\pm$ 0.5 & \textbf{0.6 $\pm$ 0.3} \\
            \protein{Arp3} &  0.8 $\pm$ 0.4 & \textbf{1.2 $\pm$ 0.3} & 2.4 $\pm$ 0.4 & 3.2 $\pm$ 0.4 & 4.2 $\pm$ 0.4 & 2.1 $\pm$ 0.5 \\
            \protein{Cki2} &  -0.2 $\pm$ 0.3 & \textbf{0.3 $\pm$ 0.5} & 1.0 $\pm$ 0.3 & 2.5 $\pm$ 0.3 & 3.6 $\pm$ 0.5 & 1.2 $\pm$ 0.3 \\
            \protein{Mkh1} &  -0.2 $\pm$ 0.4 & 0.8 $\pm$ 0.6 & \textbf{0.5 $\pm$ 0.4} & 4.6 $\pm$ 0.5 & 6.6 $\pm$ 0.5 & 2.4 $\pm$ 0.6 \\
            \protein{Sid2} &  -0.6 $\pm$ 0.3 & \textbf{0.8 $\pm$ 0.4} & 1.0 $\pm$ 0.5 & 4.5 $\pm$ 0.5 & 3.2 $\pm$ 0.6 & 1.1 $\pm$ 0.6 \\
            \protein{Tea1} & -0.1 $\pm$ 0.4  & \textbf{0.8 $\pm$ 0.5} & \textbf{0.8 $\pm$ 0.5} & 4.4 $\pm$ 0.3 & 2.8 $\pm$ 0.5 & 1.1 $\pm$ 0.4 \\
            \hline
            6 proteins &  -0.1 $\pm$ 0.2 & \textbf{0.8 $\pm$ 0.2} & 1.1 $\pm$ 0.2 & 3.7 $\pm$ 0.1 & 3.8 $\pm$ 0.2 & 1.4 $\pm$ 0.1 \\[-0.2cm]
        \end{tabular}
    \end{center}
    \caption{Results of C2ST with the WGAN-GP objective comparing several multi-channel models w.r.t.\ the real images.
        All the models were trained with WGAN-GP.
        The values in this table are directly comparable to the ones in Table~\ref{tbl:wdist:realVsReal}.\vspace{-0.3cm}
        \label{tbl:wdist:wgansVsReal}}
\end{table*}

\vspace{-4mm}\paragraph{Results.} Table~\ref{tbl:wdist:wgansVsReal} shows the results of C2ST applied to several models with multiple output channels (see Section~\ref{sec:multi-channel}): the multi-channel model and its separable version, the star-shaped model and the two baselines, which do not align green channels of different classes with the same red channel: one-class generators trained individually for each class and their separable versions. All the models were trained with WGAN-GP with the same ratio of the width of the generator tower to the number of output channels.

We observe that the individual one-class WGAN-GP models lead to higher quality compared to all the models outputting synchronized channels for all the classes.
Among the models that synchronize channels, the star-shaped model performs best, but for some proteins there is a significant drop in quality w.r.t.\ the one-class models.

\subsection{Optimization to Reconstruct the Test Set}
\label{sec:exp:optimization}
One of the common failures of GANs is the loss of modes from the distribution, usually referred to as mode collapse.
There is evidence~\cite{Poole2016} that image quality can be inversely correlated with mode coverage.
To test for the mode collapse, we perform an experiment proposed in~\cite{Metz2017unrolled}, where for a fixed trained generator~$G$ we examine how well it can reconstruct images from a held out test set. For each image in the test set, we minimize the $L_2$-distance (normalized by the number of pixels) between the generated and test images w.r.t.\ the noise vector~$\vec{z}$.  We call this task regular reconstruction.
We use 50 iterations of L-BFGS and run it 5 times to select the best reconstruction.
We also performed an additional task, separable reconstruction, which examines the ability of separable networks to reproduce modes of the green channel conditioned on the red.  In this task, we use a two-step procedure: first, we minimize the $L_2$-error between the red channels holding the green noise fixed, and then we minimize the $L_2$-error in the green channel while keeping the red noise fixed at it's optimized value.
To complete the study, we also report the negative log likelihood (NLL) w.r.t.\ the prior~$\P_{\vec{z}}$ of the noise vectors~$\vec{z}$ obtained with a reconstruction procedure.
As a baseline for the reconstruction error, we show the nearest neighbor cell (in both red and green channels) from the training set and the average $L_2$-distance to the nearest neighbors.
As a baseline for NLL, we show the averaged NLL for the random point generated from~$\P_{\vec{z}}$.

\begin{figure}
    \begin{center}
        \begin{tabular}{@{}c@{\:}c@{\:}c@{\:}c@{\:}c@{\:}c@{}}
            \includegraphics[trim = 0mm 0mm 88mm 0mm, clip, width=0.16\columnwidth]{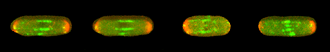}
            &
            \includegraphics[trim = 87.5mm 0mm 0.5mm 0mm, clip, width=0.16\columnwidth]{figures/reconstructions/wgangp-indep/Alp14_batch0_image26.png}
            &
            \includegraphics[trim = 29.5mm 0mm 58.5mm 0mm, clip, width=0.16\columnwidth]{figures/reconstructions/wgangp-indep/Alp14_batch0_image26.png}
            &
            \includegraphics[trim = 29.5mm 0mm 58.5mm 0mm, clip, width=0.16\columnwidth]{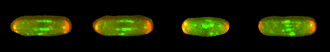}
            &
            \includegraphics[trim = 29.5mm 0mm 58.5mm 0mm, clip, width=0.16\columnwidth]{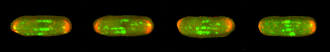}
            &
            \includegraphics[trim = 29.5mm 0mm 58.5mm 0mm, clip, width=0.16\columnwidth]{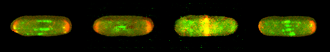} \\[-0.5mm]
            \includegraphics[trim = 0mm 0mm 88mm 0mm, clip, width=0.16\columnwidth]{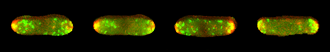}
            &
            \includegraphics[trim = 87.5mm 0mm 0.5mm 0mm, clip, width=0.16\columnwidth]{figures/reconstructions/wgangp-indep/Arp3_batch0_image3.png}
            &
            \includegraphics[trim = 29.5mm 0mm 58.5mm 0mm, clip, width=0.16\columnwidth]{figures/reconstructions/wgangp-indep/Arp3_batch0_image3.png}
            &
            \includegraphics[trim = 29.5mm 0mm 58.5mm 0mm, clip, width=0.16\columnwidth]{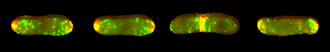}
            &
            \includegraphics[trim = 29.5mm 0mm 58.5mm 0mm, clip, width=0.16\columnwidth]{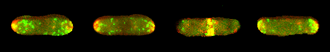}
            &
            \includegraphics[trim = 29.5mm 0mm 58.5mm 0mm, clip, width=0.16\columnwidth]{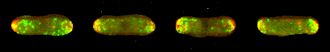} \\[-0.5mm]
            \includegraphics[trim = 0mm 0mm 88mm 0mm, clip, width=0.16\columnwidth]{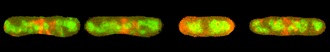}
            &
            \includegraphics[trim = 87.5mm 0mm 0.5mm 0mm, clip, width=0.16\columnwidth]{figures/reconstructions/wgangp-indep/Cki2_batch0_image5.png}
            &
            \includegraphics[trim = 29.5mm 0mm 58.5mm 0mm, clip, width=0.16\columnwidth]{figures/reconstructions/wgangp-indep/Cki2_batch0_image5.png}
            &
            \includegraphics[trim = 29.5mm 0mm 58.5mm 0mm, clip, width=0.16\columnwidth]{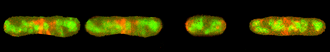}
            &
            \includegraphics[trim = 29.5mm 0mm 58.5mm 0mm, clip, width=0.16\columnwidth]{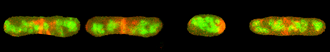}
            &
            \includegraphics[trim = 29.5mm 0mm 58.5mm 0mm, clip, width=0.16\columnwidth]{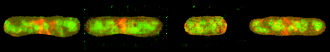} \\[-0.5mm]
            \includegraphics[trim = 0mm 0mm 88mm 0mm, clip, width=0.16\columnwidth]{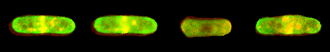}
            &
            \includegraphics[trim = 87.5mm 0mm 0.5mm 0mm, clip, width=0.16\columnwidth]{figures/reconstructions/wgangp-indep/Mkh1_batch0_image1.png}
            &
            \includegraphics[trim = 29.5mm 0mm 58.5mm 0mm, clip, width=0.16\columnwidth]{figures/reconstructions/wgangp-indep/Mkh1_batch0_image1.png}
            &
            \includegraphics[trim = 29.5mm 0mm 58.5mm 0mm, clip, width=0.16\columnwidth]{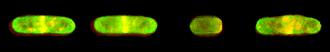}
            &
            \includegraphics[trim = 29.5mm 0mm 58.5mm 0mm, clip, width=0.16\columnwidth]{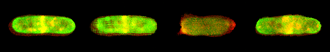}
            &
            \includegraphics[trim = 29.5mm 0mm 58.5mm 0mm, clip, width=0.16\columnwidth]{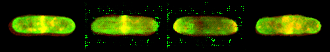} \\[-1mm]
            {\small (a)} &  {\small (b)} & {\small (c)} & {\small (d)} & {\small (e)} & {\small (f)} \\[-2mm]
        \end{tabular}
    \end{center}
    \caption{Examples of cell reconstructions.
        (a)~-- a test image; (b)~-- the $L_2$ nearest neighbor; (c)~-- regular reconstruction by one-class separable WGAN-GP; (d)~-- regular reconstruction by star-shaped WGAN-GP; (e)~-- separable reconstruction by star-shaped WGAN-GP; (f)~-- separable reconstruction by star-shaped GAN.
        An extended version of this figure is given in Appendix~\ref{app:reconstructionPlots}.
        \vspace{-2mm}
        \label{fig:reconstructionExamples}}
\end{figure}

\begin{figure}[t]
    \centering
    \begin{tabular}{c@{}c}
        \includegraphics[width=0.48\columnwidth, trim = 2mm 0mm 1.5mm 2mm, clip]{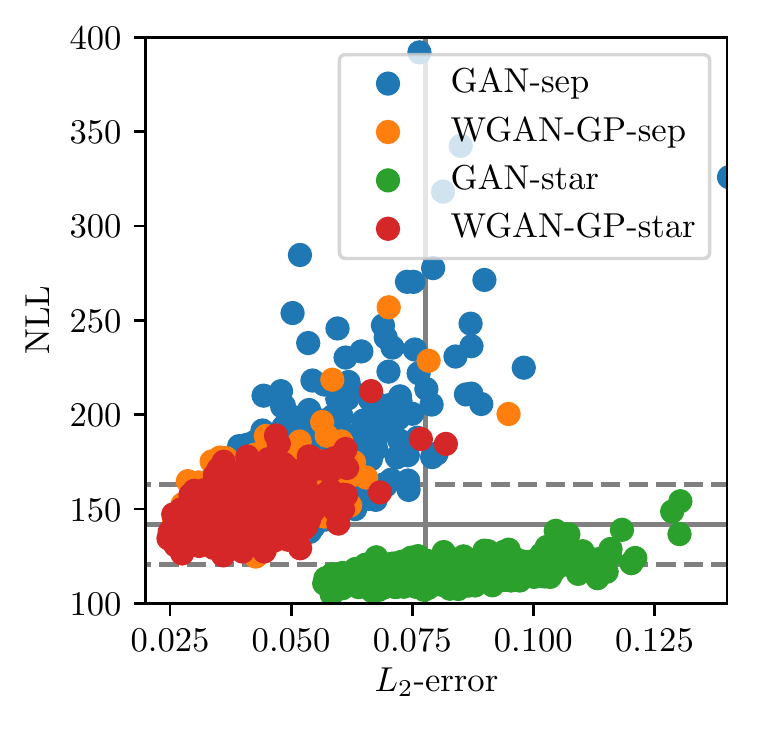} 
        &
        \includegraphics[width=0.48\columnwidth, trim = 0mm 0mm 3.5mm 2mm, clip]{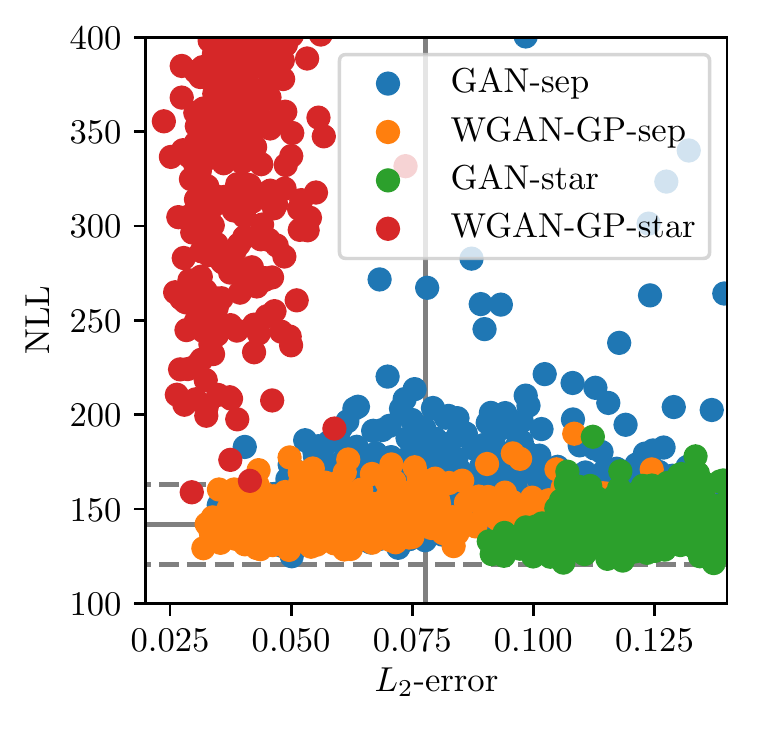}\\[-0.3cm]
        {\small (a) regular reconstruction} & {\small (b) separable reconstruction}\\[1mm]
    \end{tabular}
    \caption{Reconstruction errors against negative log likelihood (NLL) of  the latent vectors found by reconstruction.
        We show all the cells corresponding to protein \protein{Mkh1}, which appears to be the hardest for the star-shaped models.
        The vertical gray line shows the median $L_2$-error of the nearest neighbor.
        Horizontal gray lines show mean NLL ($\pm$ 3 std) of the noise sampled from the Gaussian prior.
        In the separable (red-first) setting, the star-shaped model trained with GAN provides very bad reconstructions, whereas the same model trained with WGAN-GP results in high NLL values.
        An extended version of this figure is given in Appendix~\ref{app:reconstructionPlots}.
        \vspace{-2mm}
        \label{fig:reconstruction_Mkh1}}
\end{figure}

We apply the reconstruction procedure to evaluate four models: separable one-class and star-shaped models trained with both GAN and WGAN-GP algorithms.
Figure~\ref{fig:reconstructionExamples} and Table~\ref{tbl:reconstruction} present qualitative and quantitative results, respectively.
For all the measurements, we report the median values and the median absolute deviation.
In Figure~\ref{fig:reconstruction_Mkh1}, we plot reconstruction errors vs. NLL values for the \protein{Mkh1}, which was the hardest protein in the separable reconstruction task.

Analyzing the results, we observe that separable reconstruction is a harder task than the single step procedure. Second, WGAN-GP models can reconstruct better, probably because they suffer less from the mode collapse.
And finally, the star-shaped models do not degrade the performance in terms of reconstruction, except for some hard proteins (see more details in Appendix~\ref{app:reconstructionPlots}).

\begin{table}
    \centering
    \begin{tabular}{@{}l@{\:}|@{\;}l|ll@{}}
        & &      $L_2$-error &      NLL \\
        \hline 
        & Nearest neighbors & $0.079\pm0.009$ & - \\
        & Gaussian noise & - & $142\pm5$ \\
        \hline
        \multirow{4}{*}{\rotatebox[origin=c]{90}{regular}} & GAN-sep &  $0.053\pm0.007$ &  $166\pm17$ \\
        & WGAN-GP-sep     &  $0.043\pm0.006$ &  $149\pm8$ \\
        & GAN-star      &  $0.061\pm0.008$ &  $\mathbf{139\pm12}$ \\
        & WGAN-GP-star  &  $\mathbf{0.041\pm0.005}$ &  $150\pm8$ \\
        \hline
        \multirow{4}{*}{\rotatebox[origin=c]{90}{separable}} & GAN-sep         &  $0.069\pm0.011$ &  $158\pm13$ \\
        & WGAN-GP-sep     &  $0.062\pm0.009$ &  $143\pm6$ \\
        & GAN-star      &  $0.074\pm0.011$ &  $\mathbf{142\pm7}$ \\
        & WGAN-GP-star  &  $\mathbf{0.058\pm0.010}$ &  $143\pm7$
    \end{tabular}
    \vspace{2mm}
    \caption{ Reconstruction experiment. For the four trained models (GAN/WGAN-GP and separable one-class/star-shaped), we report $L_2$-errors of the reconstructions and the negative log likelihoods (NLL) of the latent vectors found by the reconstruction.\vspace{-6mm}
        \label{tbl:reconstruction}}
\end{table}

\subsection{Progression Through the Cell Cycle}
\label{sec:cellCycle}
As described in Section \ref{sec:biobackground}, the localization of \protein{Bgs4} can be used to accurately pinpoint the cell cycle stage. However, not nearly as much as is known about how the localization of the other proteins changes within the cell cycle~\cite{martin2014cell}.

\begin{figure}
    \begin{tabular}{@{}m{0.05\columnwidth}@{\!\!\!\!\!}m{0.95\columnwidth}}
        &
        \parbox{0.95\columnwidth}{
            \hspace{0.3cm}
            \protein{Bgs4}
            \hspace{0.1mm}
            \protein{Alp14}
            \hspace{0.1mm}
            \protein{Arp3}
            \hspace{0.6mm}
            \protein{Cki2}
            \hspace{1mm}
            \protein{Mkh1}
            \hspace{1mm}
            \protein{Sid2}
            \hspace{1mm}
            \protein{Tea1}} \\
        $\xdownarrow{20mm}$ &
        \includegraphics[width=0.98\columnwidth]{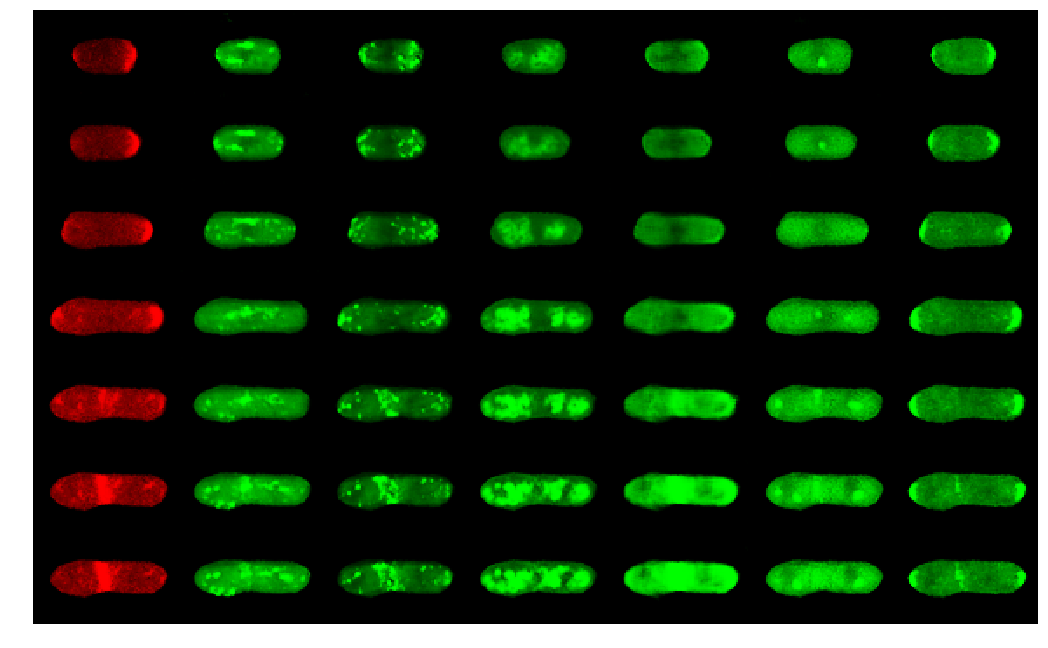}\\[-1mm]
    \end{tabular}
    \caption{Cell cycle of a star-shaped WGAN-GP model. \label{fig:cellCycle}}
    \vspace{-5mm} 
\end{figure}

Using our separable GAN architecture, we can interpolate between points in the latent space~\cite{White2016slerp} to move across the different stages of growth and division. Due to the architecture of our network, the output of the green channel will always remain consistent with the red output.
We show an example of the reconstructed cell cycle in Figure~\ref{fig:cellCycle} and several animated examples in the Suppl.\ Mat.\ \cite{Osokin2017supplmat}.  As a validation of our approach, \protein{Arp3} is seen gradually moving a dot like pattern at the tips of the cell towards the middle of the cell during mitosis, as has been previously described in the literature~\cite{yan2012meiotic}.

It's important to highlight that the LIN dataset lacks true multi-channel (3+) images, and as such, we are unable to compare how our generated multi-channel images compare to real fluorescent images.  We hope that as more datasets in biology become open, we will have a better baseline to compare our model too.

\vspace{-1mm}\section{Conclusion}
\label{sec:conclusion}

\vspace{-1mm}Although generative modeling has seen an explosion in popularity in the last couple of years, so far it has mostly been applied to the synthesis of real world images.  Our results in this paper suggest that modern generative models can be fruitfully applied to images obtained by fluorescent microscopy.  By leveraging correlation between different image channels, we were able to simulate the localization of multiple proteins throughout the cell cycle.  This could enable in the future the exploration of uninvestigated, inaccessible or unaffordable biological/biomedical experiments, to catalyze new discoveries and potentially enable new diagnostic and prognostic bioimaging applications. 

\vspace{3mm}\noindent {\bf Acknowledgements.}
A. Osokin was supported by the ERC grant Activia (no. 307574).  F. Vaggi was supported by a grant from the chair Havas-Dauphine.

{\small
\bibliographystyle{ieee}
\bibliography{egbib}
}

\clearpage
\appendix

\twocolumn[
    \begin{center}
        {\Large \bf Supplementary materials for ``GANs for Biological Image Synthesis''}
    \end{center}
]

\section{Comparison of C2ST variants\label{app:c2st}}
In this section, we report an experiment comparing the behavior of several variants of classifier two-samples test (C2ST) based on different ways of training the classifier.
We consider three approaches to train the classier used in C2ST, which come from GAN~\cite{Goodfellow2014gan,Radford2016dcgan}, WGAN~\cite{Arjovsky2017wgan} and WGAN-GP~\cite{Gulrajani2017wgangp}.
In the case of GAN, we simply train a classifier using the negative cross-entropy loss and report the negative loss on the test-test set and freshly generated images as the C2ST score (note that such score is always negative).
In the case of WGAN, we train a classifier with all the weights clipped to the segment~$[-0.01, 0.01]$ and use the negation of~\eqref{eq:wasserDual} as the C2ST score (note that such scores are always non-negative).
In the case of WGAN-GP, we train a classifier with the regularizer based on the~$L_2$-penalty on the gradient norm~\eqref{eq:wassesPractical} (with the regularizer weight equal to 10) and use the negation of~\eqref{eq:wassesPractical} as the C2ST score (note that these scores can be negative when the two collections of samples are similar).

For C2ST with GAN or WGAN-GP, we use the Adam optimizer~\cite{kingma2014adam}.
For C2ST with WGAN, we use the RMSprop optimizer~\cite{Hinton2012rmsprop}.
We run all optimizers for 5000 iterations with the parameters coming from the corresponding GAN methods (see our implementation~\cite{Osokin2017supplmat} for the details).

\begin{figure}
    \begin{tabular}{@{}c@{\!\!}c@{}}
        \includegraphics[width=0.5\columnwidth]{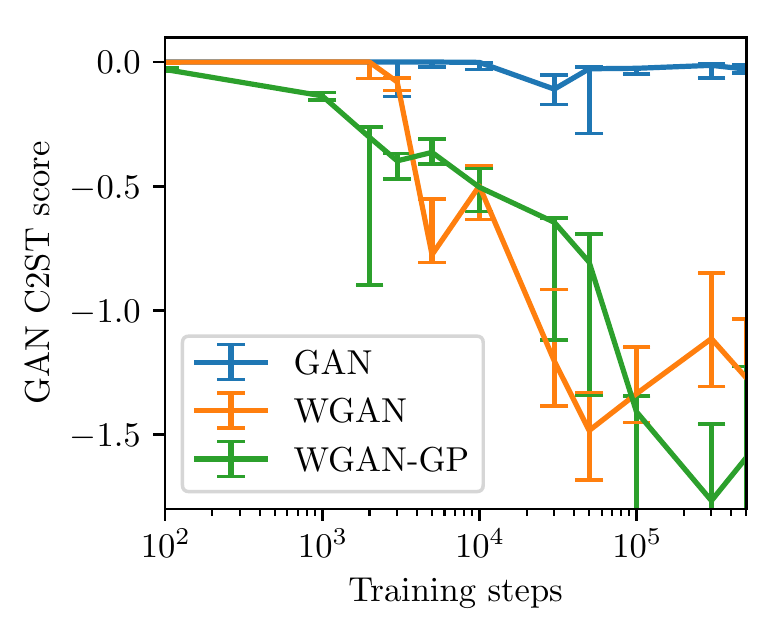}
        &
        \includegraphics[width=0.5\columnwidth]{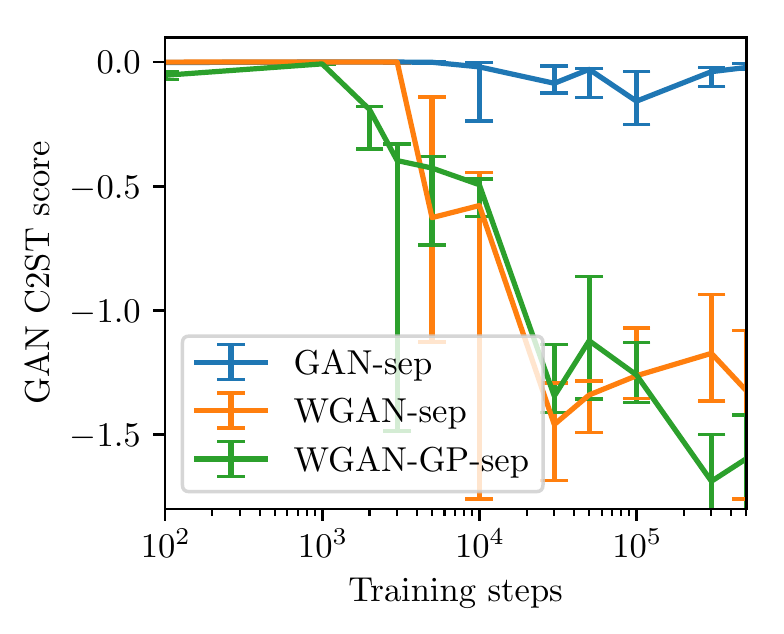}
        \\
        \parbox{0.5\columnwidth}{ \centering\small (a): non-separable models \\ GAN C2ST scores}
        &
        \parbox{0.5\columnwidth}{ \centering\small (b): separable models \\ GAN C2ST scores}
        \\[5mm]
        \includegraphics[width=0.5\columnwidth]{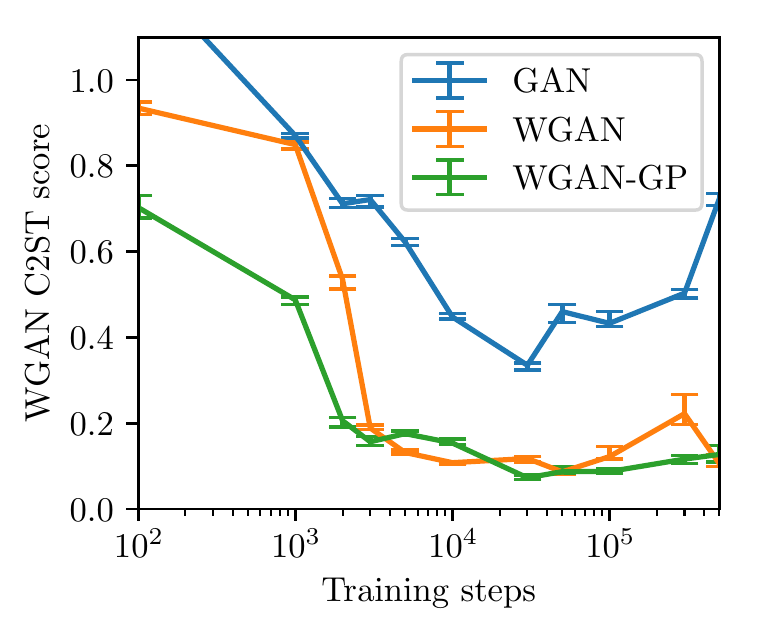}
        &
        \includegraphics[width=0.5\columnwidth]{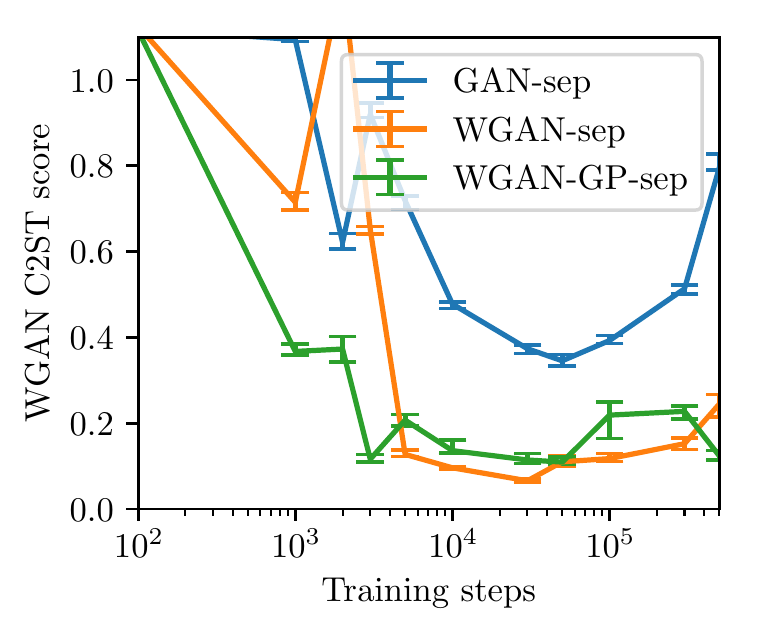}
        \\
        \parbox{0.5\columnwidth}{ \centering\small (c): non-separable models \\ WGAN C2ST scores}
        &
        \parbox{0.5\columnwidth}{ \centering\small (d): separable models \\ WGAN C2ST scores}
        \\[5mm]
        \includegraphics[width=0.5\columnwidth]{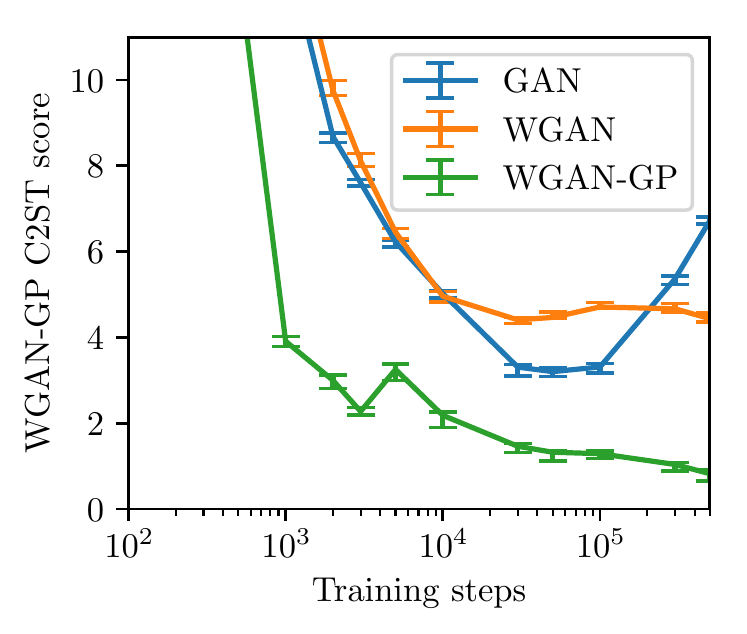}
        &
        \includegraphics[width=0.5\columnwidth]{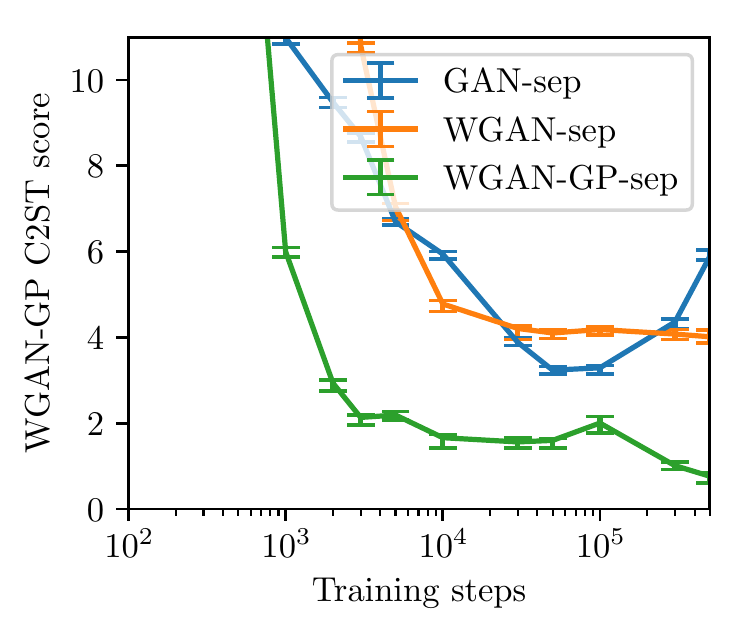}
        \\
        \parbox{0.5\columnwidth}{ \centering\small (e): non-separable models \\  WGAN-GP C2ST scores}
        &
        \parbox{0.5\columnwidth}{ \centering\small separable models \\ WGAN-GP C2ST scores}
        \\[5mm]
    \end{tabular}
    \caption{Comparison of C2ST based on different ways to train the discriminator.
        For all C2ST scores, the lower the better.
        For real images, GAN gives~$-3.7\pm 0.7$, WGAN gives $0.0\pm 0.0$, WGAN-GP gives $-0.7\pm0.6$. \label{fig:c2stVariants}}
\end{figure}

\begin{figure*}
    \begin{center}
        \begin{tabular}{@{}m{8mm}@{\:}|@{\:}m{8mm}@{\:}|@{}m{44mm}@{}|m{8mm}@{\:}|@{}m{44mm}@{}|m{8mm}@{\:}|@{}m{44mm}@{}}
            Steps
            & C2ST & \parbox{44mm}{\centering Samples of GAN}
            & C2ST & \parbox{44mm}{\centering Samples of WGAN}
            & C2ST & \parbox{44mm}{\centering Samples of WGAN-GP}
            \\[0mm]
            \hline
            100 & \parbox{8mm}{\small $97.0\pm$ \\[-0.5mm] $0.2$} &
            \vspace{0.5mm}\includegraphics[width=44mm]{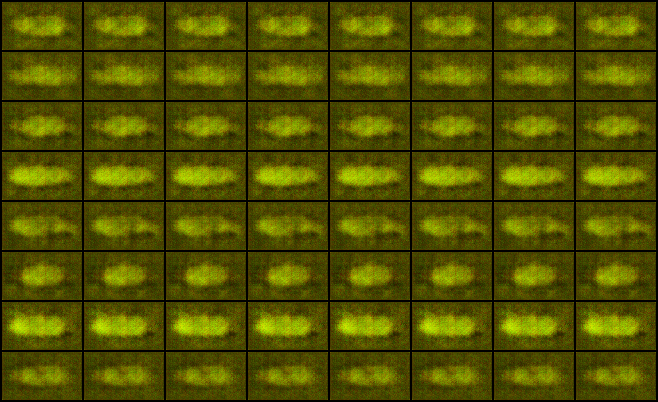}
            &
            \parbox{8mm}{\small $147\pm$ \\[-0.5mm] $0.2$} &
            \vspace{0.5mm}\includegraphics[width=44mm]{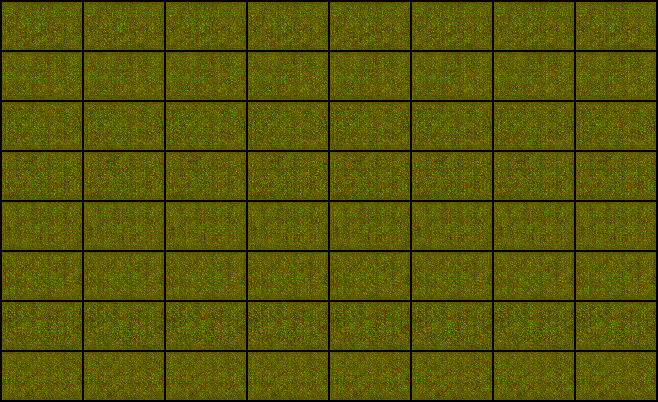}
            &
            \parbox{8mm}{\small $32.8\pm$ \\[-0.5mm] $0.2$} &
            \vspace{0.5mm}\includegraphics[width=44mm]{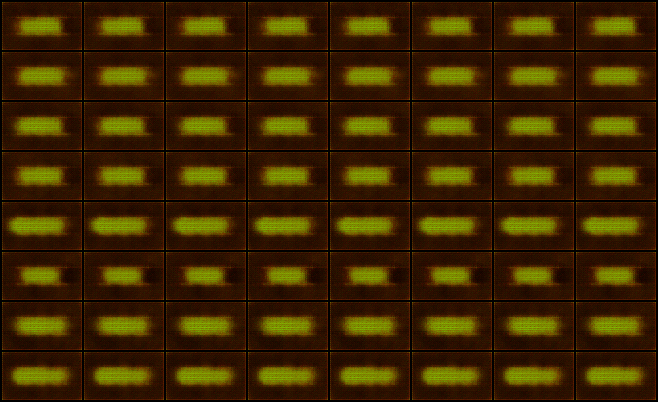}
            \\[-1.3mm]
            \hline
            1k & \parbox{8mm}{\small $11.0\pm$ \\[-0.5mm] $0.1$} &
            \includegraphics[width=44mm]{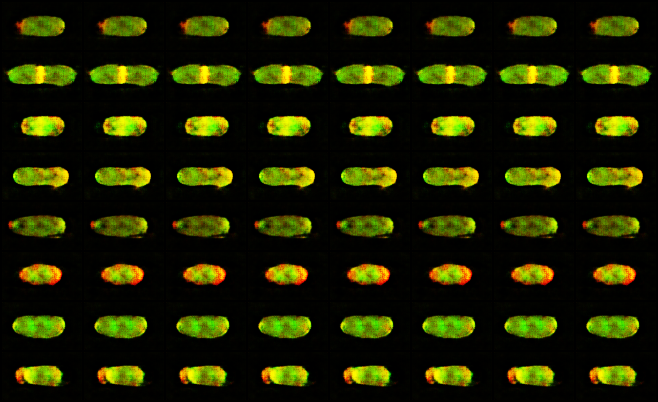}
            &
            \parbox{8mm}{\small $57.1\pm$ \\[-0.5mm] $0.3$} &
            \includegraphics[width=44mm]{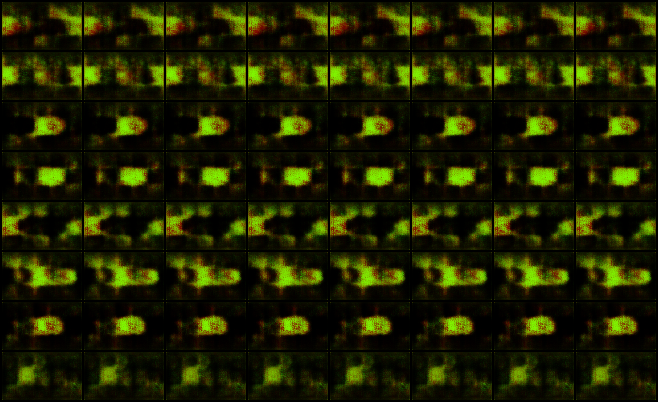}
            &
            \parbox{8mm}{\small $6.0\pm$ \\[-0.5mm] $0.1$} &
            \includegraphics[width=44mm]{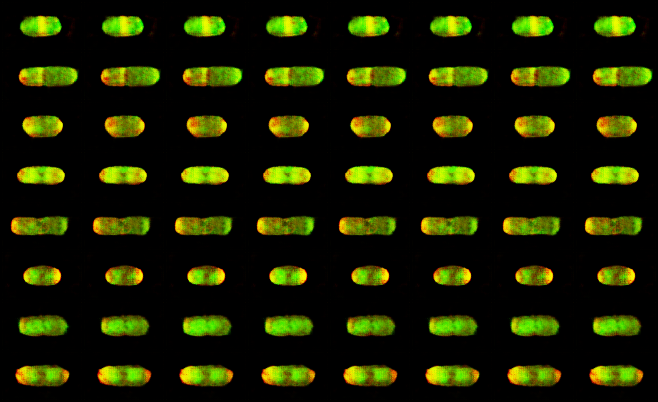}
            \\[-1.3mm]
            \hline
            2k & \parbox{8mm}{\small $9.5\pm$ \\[-0.5mm] $0.1$} &
            \includegraphics[width=44mm]{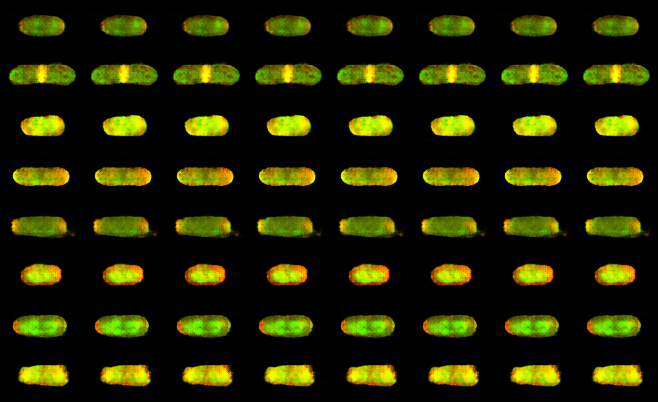}
            &
            \parbox{8mm}{\small $23.8\pm$ \\[-0.5mm] $0.1$} &
            \includegraphics[width=44mm]{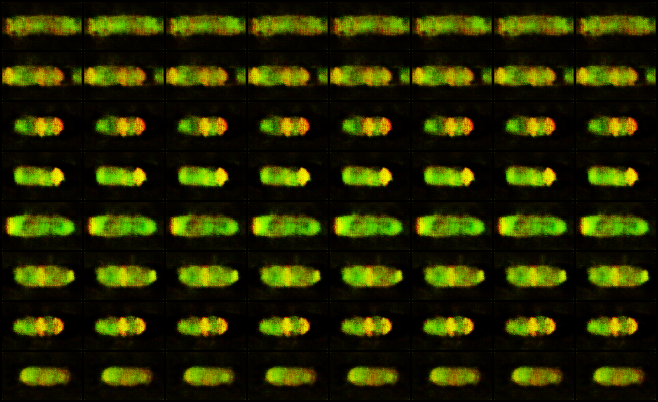}
            &
            \parbox{8mm}{\small $2.9\pm$ \\[-0.5mm] $0.1$} &
            \includegraphics[width=44mm]{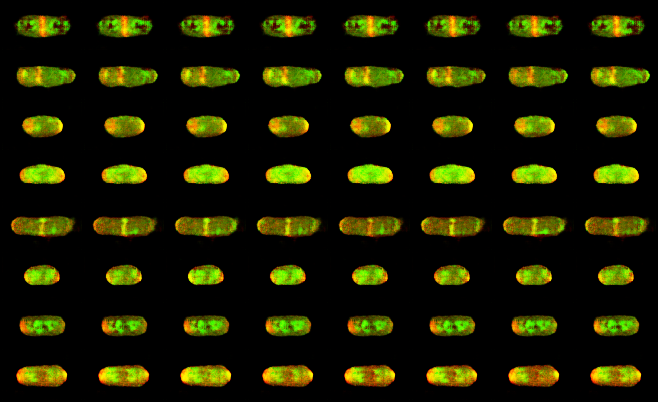}
            \\[-1.3mm]
            \hline
            5k & \parbox{8mm}{\small $6.7\pm$ \\[-0.5mm] $0.1$} &
            \includegraphics[width=44mm]{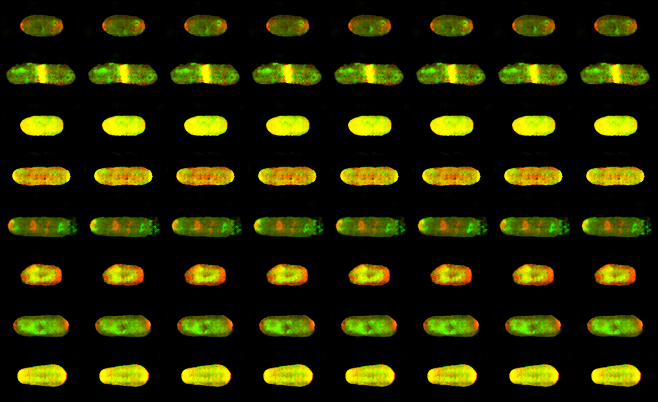}
            &
            \parbox{8mm}{\small $7.0\pm$ \\[-0.5mm] $0.1$} &
            \includegraphics[width=44mm]{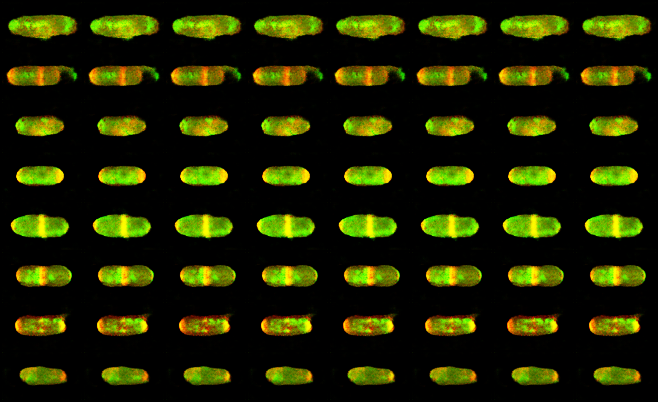}
            &
            \parbox{8mm}{\small $2.2\pm$ \\[-0.5mm] $0.1$} &
            \includegraphics[width=44mm]{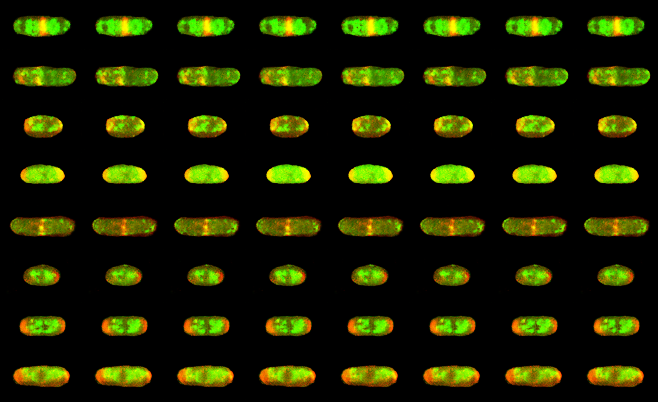}
            \\[-1.3mm]
            \hline
            10k & \parbox{8mm}{\small $5.9\pm$ \\[-0.5mm] $0.1$} &
            \includegraphics[width=44mm]{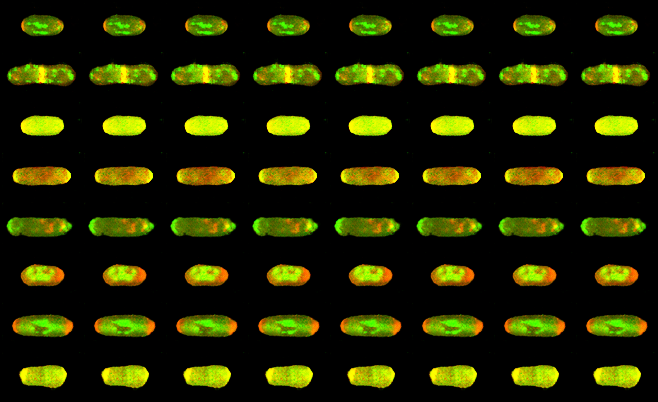}
            &
            \parbox{8mm}{\small $4.8\pm$ \\[-0.5mm] $0.1$} &
            \includegraphics[width=44mm]{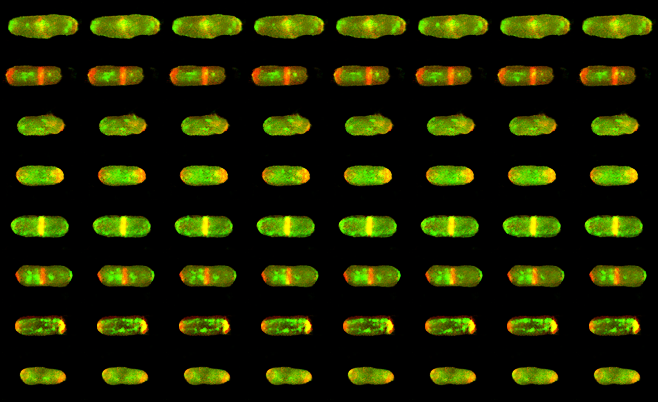}
            &
            \parbox{8mm}{\small $1.7\pm$ \\[-0.5mm] $0.1$} &
            \includegraphics[width=44mm]{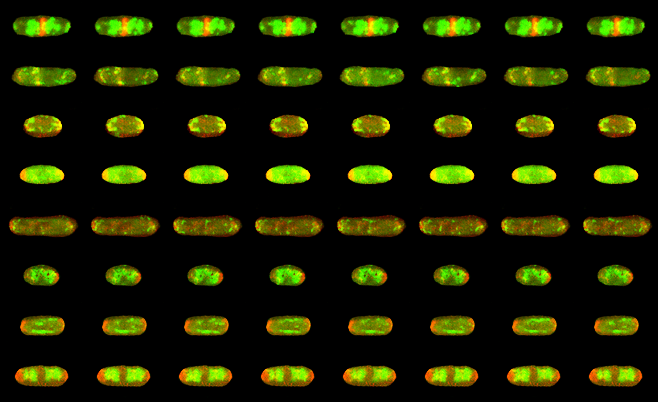}
            \\[-1.3mm]
            \hline
            50k & \parbox{8mm}{\small $3.2\pm$ \\[-0.5mm] $0.1$} &
            \includegraphics[width=44mm]{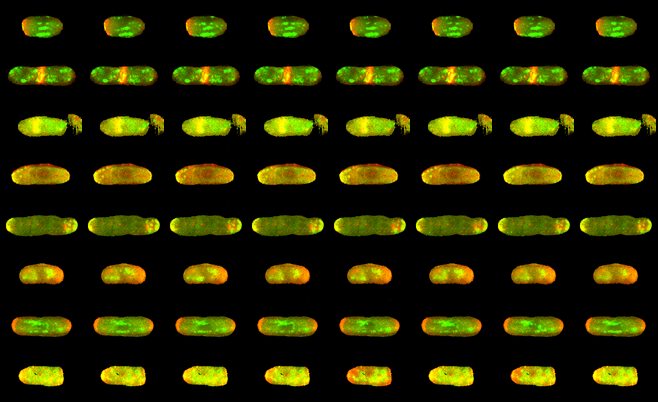}
            &
            \parbox{8mm}{\small $4.1\pm$ \\[-0.5mm] $0.1$} &
            \includegraphics[width=44mm]{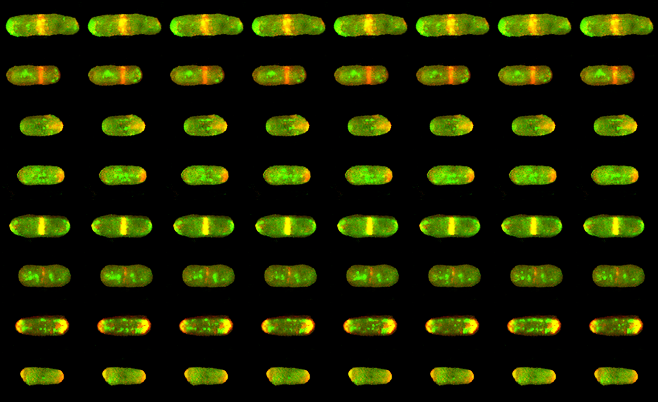}
            &
            \parbox{8mm}{\small $1.3\pm$ \\[-0.5mm] $0.1$} &
            \includegraphics[width=44mm]{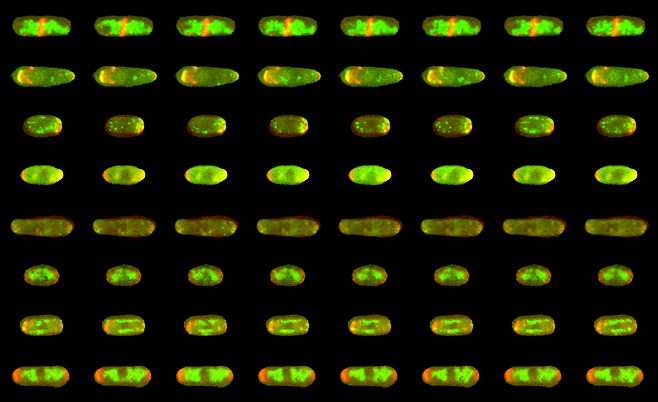}
            \\[-1.3mm]
            \hline
            500k & \parbox{8mm}{\small $5.9\pm$ \\[-0.5mm] $0.1$} &
            \includegraphics[width=44mm]{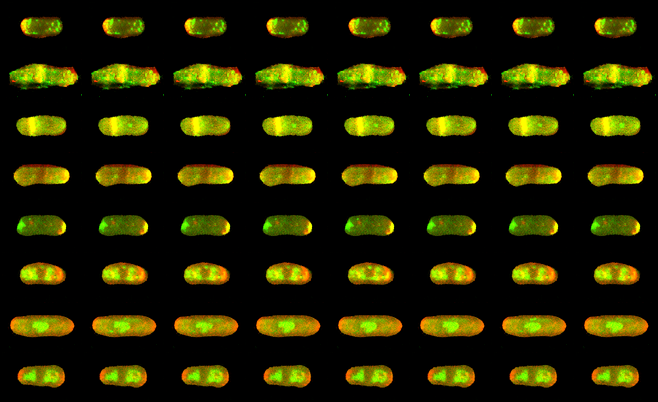}
            &
            \parbox{8mm}{\small $4.0\pm$ \\[-0.5mm] $0.1$} &
            \includegraphics[width=44mm]{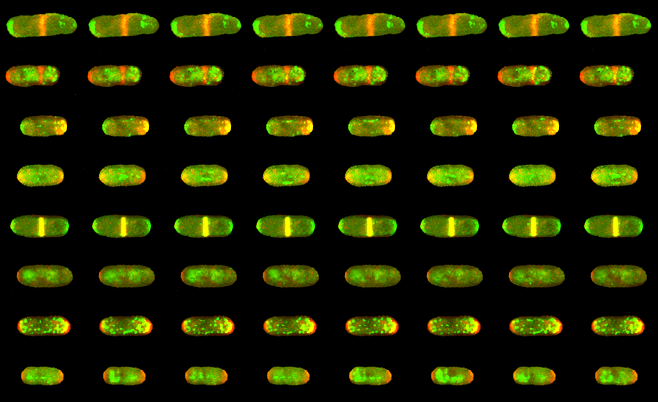}
            &
            \parbox{8mm}{\small $0.8\pm$ \\[-0.5mm] $0.1$} &
            \includegraphics[width=44mm]{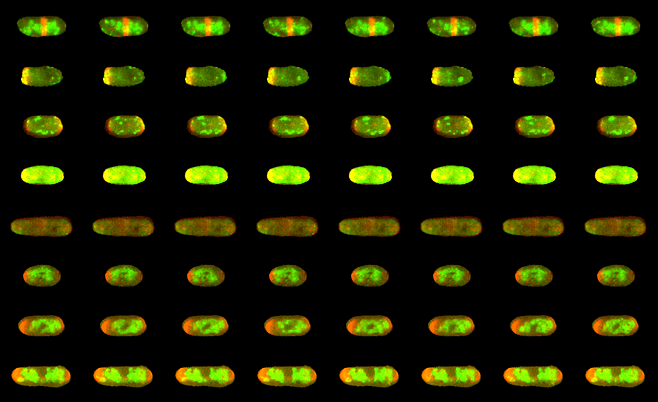}
        \end{tabular}
    \end{center}
    \vspace{-2mm}\caption{Scores of the classifier two-sample test (C2ST) between the generators and the hold-out test sets of images (an extension of Figure \ref{fig:wdistEvolution}). We report the scores of separable GAN, WGAN and WGAN-GP at different stages of training. For each line, we show the samples from the corresponding models to demonstrate that the lower C2ST scores correspond to better-looking (sharper, less artifacts, etc.) images. Best viewed in color and on a screen.\vspace{-0mm}
        \label{fig:wdistEvolutionBig}}
\end{figure*}

In Figure~\ref{fig:c2stVariants}, we apply all the three versions of C2ST after different number of training iterations of GAN, WGAN and WGAN-GP for both regular and separable (green-on-red) generators.
We repeat each measurement on 10 splits of the test set and report the median and 0.1, 0.9 quantiles.
In Figure~\ref{fig:wdistEvolutionBig}, we show samples generated by separable models trained with GAN, WGAN and WGAN-GP.
We observe that all the variants of C2ST (including the GAN version) show that the WGAN and WGAN-GP models are significantly better than the GAN ones.
This effect is likely happening due to the conditional mode collapse of the GAN models (see the first column of Figure~\ref{fig:wdistEvolutionBig}).
Comparing the C2ST score themselves, we conclude that the versions based on WGAN and WGAN-GP have less variance and are more stable.
The two latter versions perform similarly and do not show a clear winner.

\section{Reconstructing images of different proteins \label{app:reconstructionPlots}}
\begin{figure*}
    \centering
    \begin{tabular}{@{}c@{}cc@{}c@{}}
        \includegraphics[width=0.48\columnwidth, trim = 2mm 0mm 1.5mm 2mm, clip]{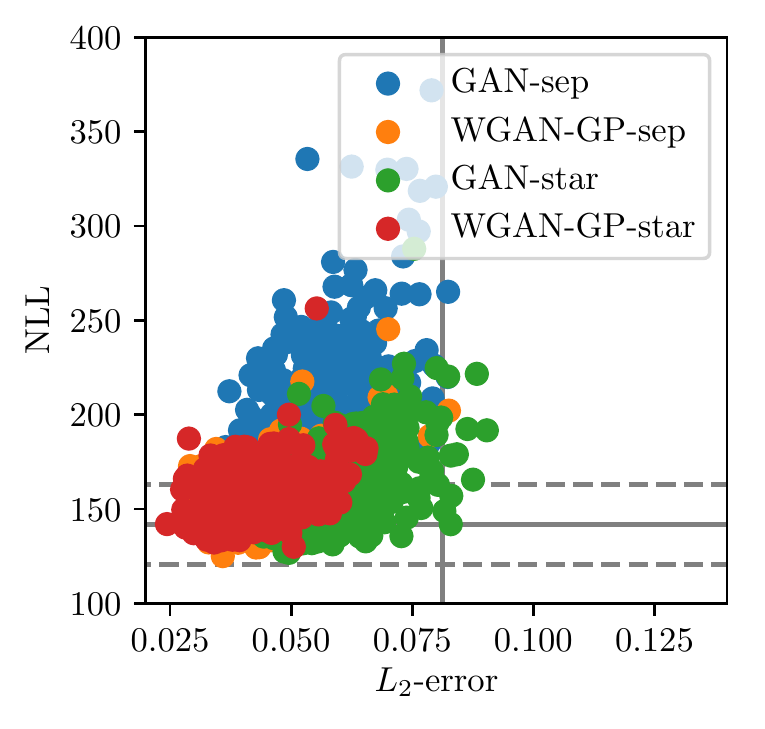} 
        &
        \includegraphics[width=0.48\columnwidth, trim = 0mm 0mm 3.5mm 2mm, clip]{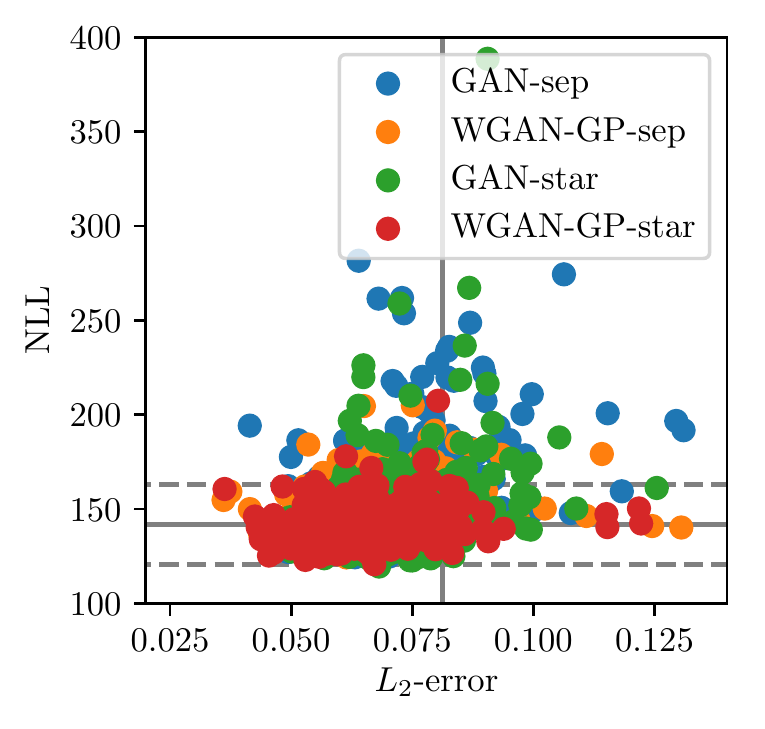}
        &
        \includegraphics[width=0.48\columnwidth, trim = 2mm 0mm 1.5mm 2mm, clip]{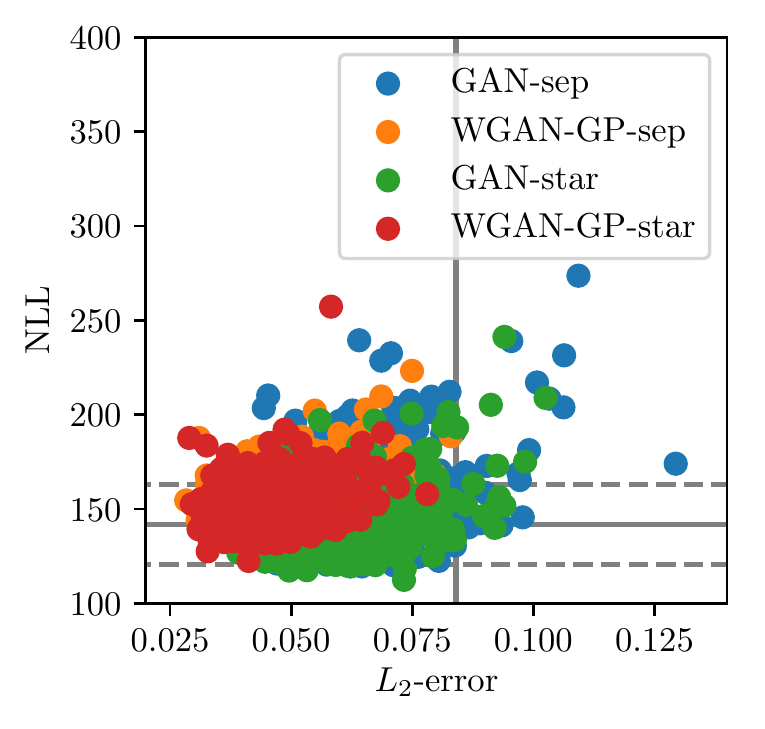} 
        &
        \includegraphics[width=0.48\columnwidth, trim = 0mm 0mm 3.5mm 2mm, clip]{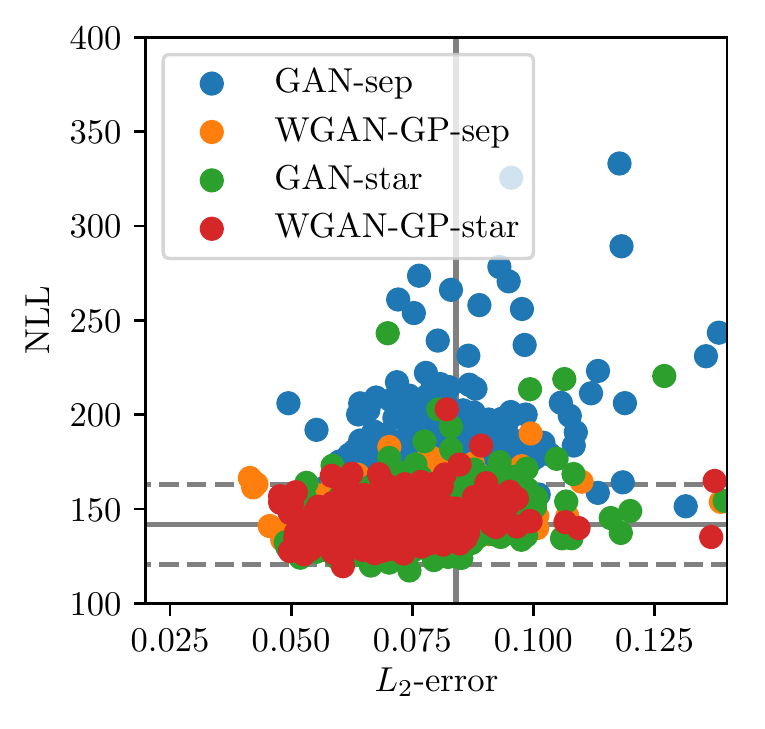}
        \\[-1mm]
        \parbox{0.5\columnwidth}{\centering\small images of \protein{Alp14} \\ regular reconstruction}
        &
        \parbox{0.5\columnwidth}{\centering\small images of \protein{Alp14} \\ separable reconstruction}
        &
        \parbox{0.5\columnwidth}{\centering\small images of \protein{Arp3} \\ regular reconstruction}
        &
        \parbox{0.5\columnwidth}{\centering\small images of \protein{Arp3} \\ separable reconstruction}
        \\[4mm]
        \includegraphics[width=0.48\columnwidth, trim = 2mm 0mm 1.5mm 2mm, clip]{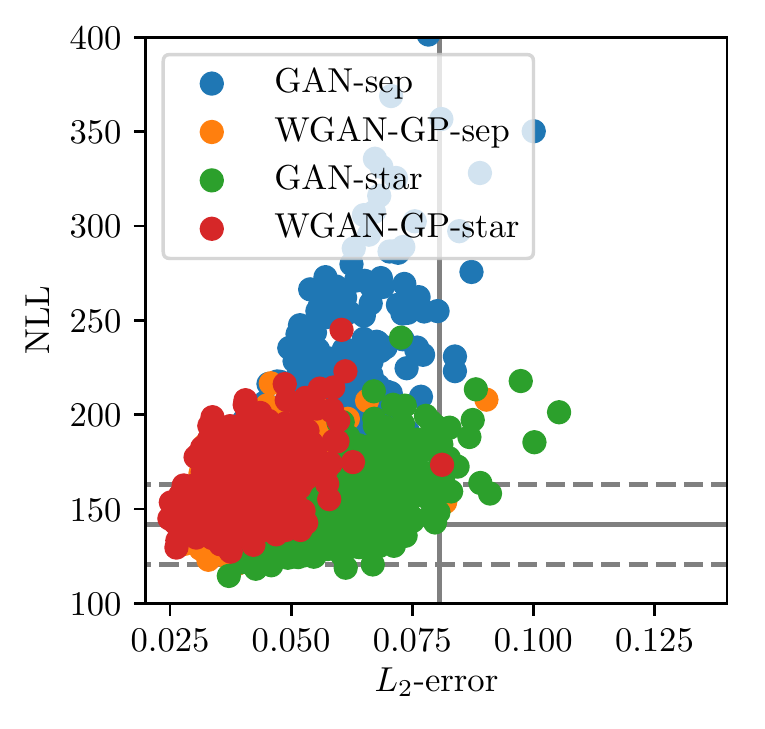} 
        &
        \includegraphics[width=0.48\columnwidth, trim = 0mm 0mm 3.5mm 2mm, clip]{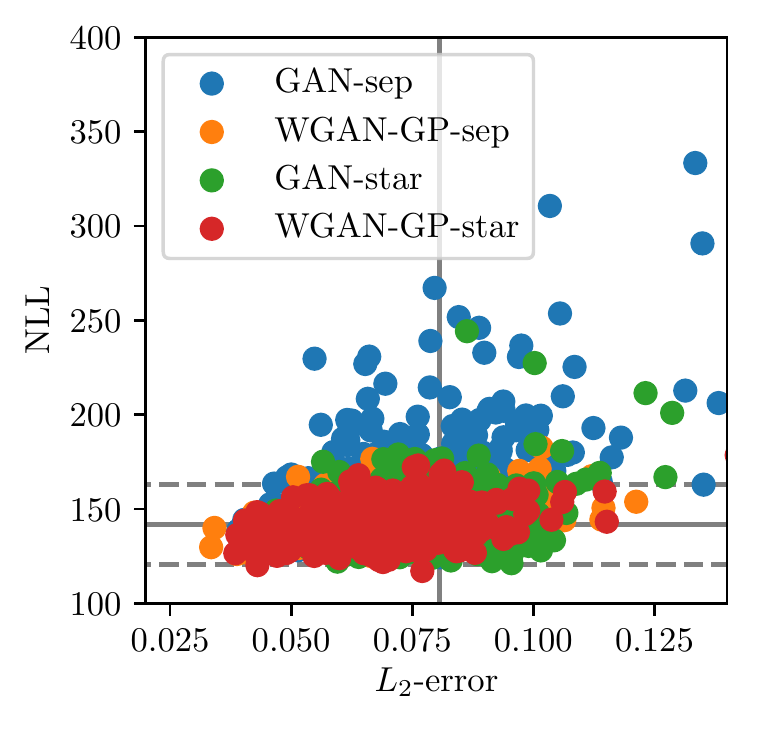}
        &
        \includegraphics[width=0.48\columnwidth, trim = 2mm 0mm 1.5mm 2mm, clip]{figures/reconstruction-regular_Mkh1.pdf} 
        &
        \includegraphics[width=0.48\columnwidth, trim = 0mm 0mm 3.5mm 2mm, clip]{figures/reconstruction-separable_Mkh1.pdf}
        \\[-1mm]
        \parbox{0.5\columnwidth}{\centering\small images of \protein{Cki2} \\ regular reconstruction}
        &
        \parbox{0.5\columnwidth}{\centering\small images of \protein{Cki2} \\ separable reconstruction}
        &
        \parbox{0.5\columnwidth}{\centering\small images of \protein{Mkh1} \\ regular reconstruction}
        &
        \parbox{0.5\columnwidth}{\centering\small images of \protein{Mkh1} \\ separable reconstruction}
        \\[4mm]
        \includegraphics[width=0.48\columnwidth, trim = 2mm 0mm 1.5mm 2mm, clip]{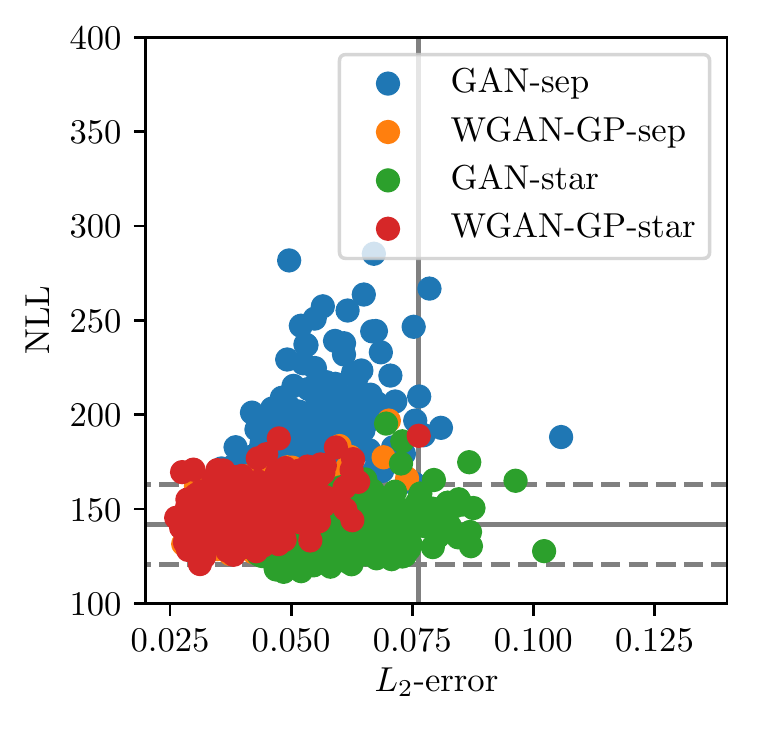} 
        &
        \includegraphics[width=0.48\columnwidth, trim = 0mm 0mm 3.5mm 2mm, clip]{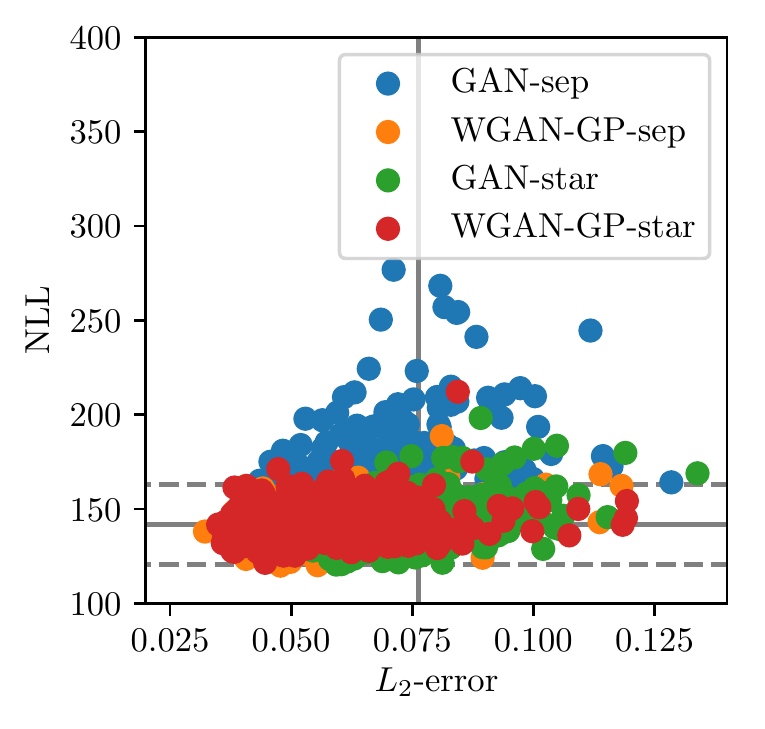}
        &
        \includegraphics[width=0.48\columnwidth, trim = 2mm 0mm 1.5mm 2mm, clip]{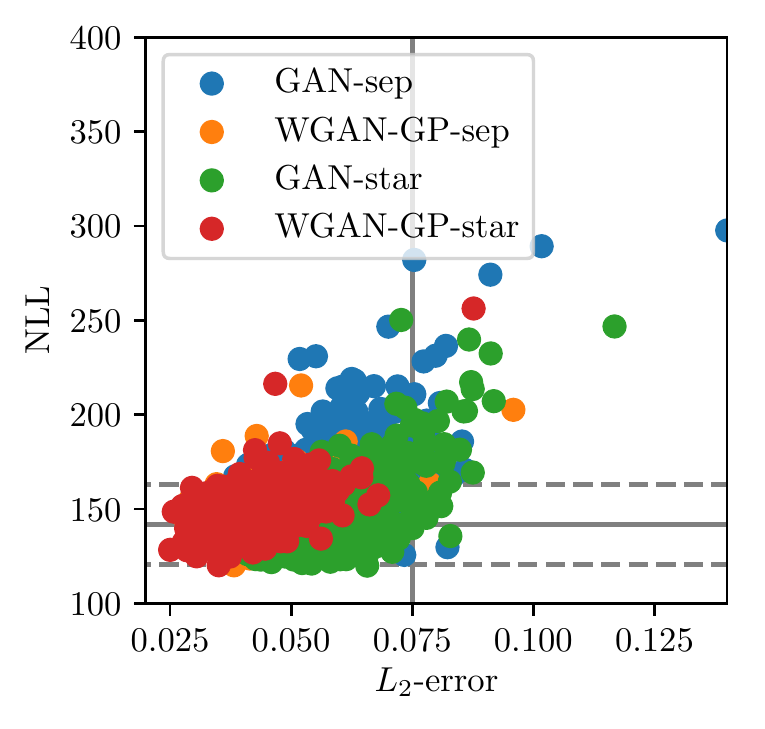} 
        &
        \includegraphics[width=0.48\columnwidth, trim = 0mm 0mm 3.5mm 2mm, clip]{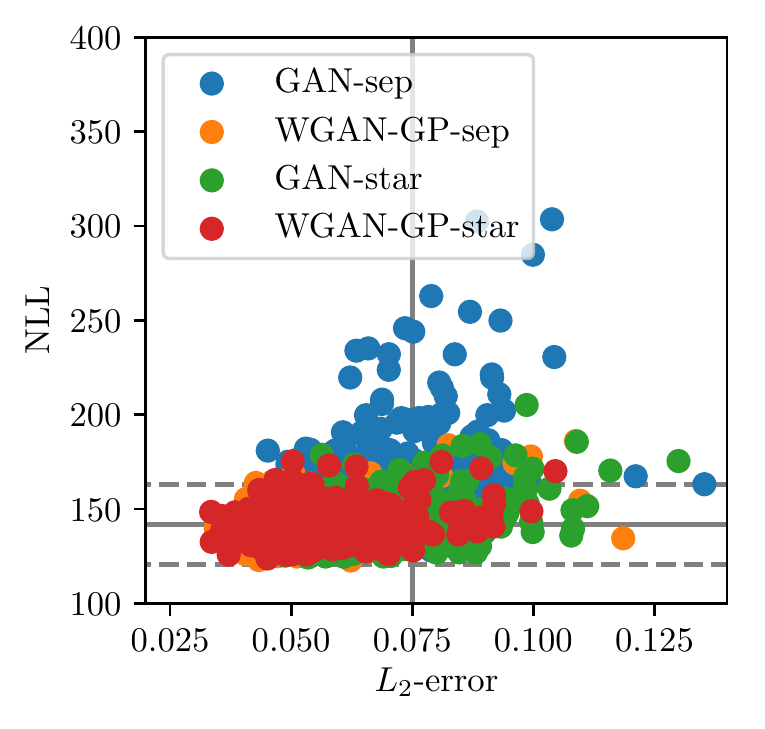}
        \\[-1mm]
        \parbox{0.5\columnwidth}{\centering\small images of \protein{Sid2} \\ regular reconstruction}
        &
        \parbox{0.5\columnwidth}{\centering\small images of \protein{Sid2} \\ separable reconstruction}
        &
        \parbox{0.5\columnwidth}{\centering\small images of \protein{Tea1} \\ regular reconstruction}
        &
        \parbox{0.5\columnwidth}{\centering\small images of \protein{Tea1} \\ separable reconstruction}
        \\[4mm]
    \end{tabular}
    \caption{Reconstruction errors against negative log likelihood (NLL) of  the latent vectors found by reconstruction (complete version of Figure~\ref{fig:reconstruction_Mkh1}).
        We show all the cells corresponding to all selected proteins.
        The vertical gray line shows the median $L_2$-error of the nearest neighbor.
        Horizontal gray lines show mean NLL ($\pm$ 3 std) of the noise sampled from the Gaussian prior.
        We observer that the images of~\protein{Mkh1} are that hardest to reconstruct for the star-shaped models.
        In the separable (red-first) setting, the star-shaped model trained with GAN provides very bad reconstructions, whereas the same model trained with WGAN-GP results in high NLL values.
        For all the other proteins, the star-shaped models can reconstruct as well as simpler separable models.
        We also conclude that the models trained with WGAN-GP reconstruct consistently better (smaller both NLL and $L_2$-error) compared to the models trained with the GAN objective.
        \vspace{-1mm}
        \label{fig:reconstruction_all}}
\end{figure*}

\begin{figure*}
    \begin{center}
        \begin{tabular}{c@{\hspace{4mm}}c}
        \begin{tabular}{@{}c@{\:}c@{\:}c@{\:}c@{\:}c@{\:}c@{}}
            \includegraphics[trim = 0mm 0mm 88mm 0mm, clip, width=0.16\columnwidth]{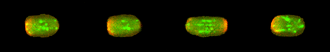}
            &
            \includegraphics[trim = 87.5mm 0mm 0.5mm 0mm, clip, width=0.16\columnwidth]{figures/reconstructions/wgangp-indep/Alp14_batch0_image0.png}
            &
            \includegraphics[trim = 29.5mm 0mm 58.5mm 0mm, clip, width=0.16\columnwidth]{figures/reconstructions/wgangp-indep/Alp14_batch0_image0.png}
            &
            \includegraphics[trim = 29.5mm 0mm 58.5mm 0mm, clip, width=0.16\columnwidth]{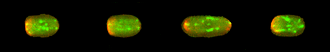}
            &
            \includegraphics[trim = 29.5mm 0mm 58.5mm 0mm, clip, width=0.16\columnwidth]{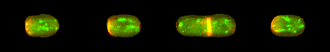}
            &
            \includegraphics[trim = 29.5mm 0mm 58.5mm 0mm, clip, width=0.16\columnwidth]{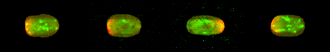} \\[-0.5mm]
            \includegraphics[trim = 0mm 0mm 88mm 0mm, clip, width=0.16\columnwidth]{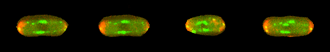}
            &
            \includegraphics[trim = 87.5mm 0mm 0.5mm 0mm, clip, width=0.16\columnwidth]{figures/reconstructions/wgangp-indep/Alp14_batch0_image1.png}
            &
            \includegraphics[trim = 29.5mm 0mm 58.5mm 0mm, clip, width=0.16\columnwidth]{figures/reconstructions/wgangp-indep/Alp14_batch0_image1.png}
            &
            \includegraphics[trim = 29.5mm 0mm 58.5mm 0mm, clip, width=0.16\columnwidth]{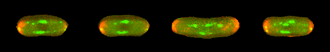}
            &
            \includegraphics[trim = 29.5mm 0mm 58.5mm 0mm, clip, width=0.16\columnwidth]{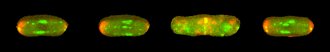}
            &
            \includegraphics[trim = 29.5mm 0mm 58.5mm 0mm, clip, width=0.16\columnwidth]{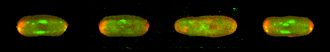} \\[-0.5mm]
            \includegraphics[trim = 0mm 0mm 88mm 0mm, clip, width=0.16\columnwidth]{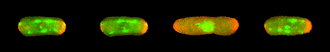}
            &
            \includegraphics[trim = 87.5mm 0mm 0.5mm 0mm, clip, width=0.16\columnwidth]{figures/reconstructions/wgangp-indep/Alp14_batch0_image10.png}
            &
            \includegraphics[trim = 29.5mm 0mm 58.5mm 0mm, clip, width=0.16\columnwidth]{figures/reconstructions/wgangp-indep/Alp14_batch0_image10.png}
            &
            \includegraphics[trim = 29.5mm 0mm 58.5mm 0mm, clip, width=0.16\columnwidth]{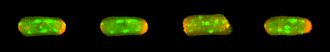}
            &
            \includegraphics[trim = 29.5mm 0mm 58.5mm 0mm, clip, width=0.16\columnwidth]{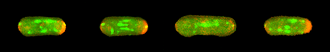}
            &
            \includegraphics[trim = 29.5mm 0mm 58.5mm 0mm, clip, width=0.16\columnwidth]{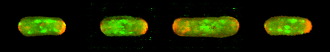} 
            \\[-0.5mm]
            \includegraphics[trim = 0mm 0mm 88mm 0mm, clip, width=0.16\columnwidth]{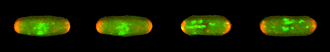}
            &
            \includegraphics[trim = 87.5mm 0mm 0.5mm 0mm, clip, width=0.16\columnwidth]{figures/reconstructions/wgangp-indep/Alp14_batch0_image11.png}
            &
            \includegraphics[trim = 29.5mm 0mm 58.5mm 0mm, clip, width=0.16\columnwidth]{figures/reconstructions/wgangp-indep/Alp14_batch0_image11.png}
            &
            \includegraphics[trim = 29.5mm 0mm 58.5mm 0mm, clip, width=0.16\columnwidth]{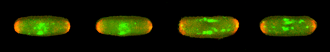}
            &
            \includegraphics[trim = 29.5mm 0mm 58.5mm 0mm, clip, width=0.16\columnwidth]{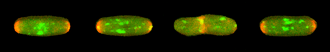}
            &
            \includegraphics[trim = 29.5mm 0mm 58.5mm 0mm, clip, width=0.16\columnwidth]{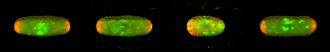} 
            \\[-0.5mm]
            \includegraphics[trim = 0mm 0mm 88mm 0mm, clip, width=0.16\columnwidth]{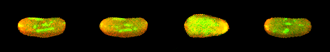}
            &
            \includegraphics[trim = 87.5mm 0mm 0.5mm 0mm, clip, width=0.16\columnwidth]{figures/reconstructions/wgangp-indep/Alp14_batch0_image12.png}
            &
            \includegraphics[trim = 29.5mm 0mm 58.5mm 0mm, clip, width=0.16\columnwidth]{figures/reconstructions/wgangp-indep/Alp14_batch0_image12.png}
            &
            \includegraphics[trim = 29.5mm 0mm 58.5mm 0mm, clip, width=0.16\columnwidth]{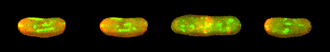}
            &
            \includegraphics[trim = 29.5mm 0mm 58.5mm 0mm, clip, width=0.16\columnwidth]{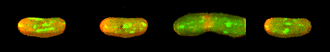}
            &
            \includegraphics[trim = 29.5mm 0mm 58.5mm 0mm, clip, width=0.16\columnwidth]{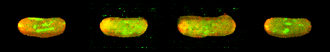} 
            \\[-0.5mm]
            \includegraphics[trim = 0mm 0mm 88mm 0mm, clip, width=0.16\columnwidth]{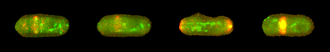}
            &
            \includegraphics[trim = 87.5mm 0mm 0.5mm 0mm, clip, width=0.16\columnwidth]{figures/reconstructions/wgangp-indep/Alp14_batch0_image13.png}
            &
            \includegraphics[trim = 29.5mm 0mm 58.5mm 0mm, clip, width=0.16\columnwidth]{figures/reconstructions/wgangp-indep/Alp14_batch0_image13.png}
            &
            \includegraphics[trim = 29.5mm 0mm 58.5mm 0mm, clip, width=0.16\columnwidth]{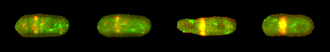}
            &
            \includegraphics[trim = 29.5mm 0mm 58.5mm 0mm, clip, width=0.16\columnwidth]{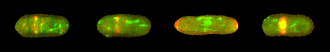}
            &
            \includegraphics[trim = 29.5mm 0mm 58.5mm 0mm, clip, width=0.16\columnwidth]{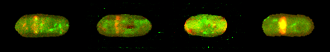}
            \\[-0.5mm]
            \includegraphics[trim = 0mm 0mm 88mm 0mm, clip, width=0.16\columnwidth]{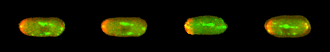}
            &
            \includegraphics[trim = 87.5mm 0mm 0.5mm 0mm, clip, width=0.16\columnwidth]{figures/reconstructions/wgangp-indep/Alp14_batch0_image14.png}
            &
            \includegraphics[trim = 29.5mm 0mm 58.5mm 0mm, clip, width=0.16\columnwidth]{figures/reconstructions/wgangp-indep/Alp14_batch0_image14.png}
            &
            \includegraphics[trim = 29.5mm 0mm 58.5mm 0mm, clip, width=0.16\columnwidth]{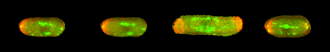}
            &
            \includegraphics[trim = 29.5mm 0mm 58.5mm 0mm, clip, width=0.16\columnwidth]{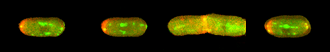}
            &
            \includegraphics[trim = 29.5mm 0mm 58.5mm 0mm, clip, width=0.16\columnwidth]{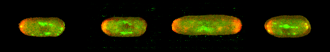} 
            \\[2mm]
            \includegraphics[trim = 0mm 0mm 88mm 0mm, clip, width=0.16\columnwidth]{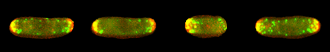}
            &
            \includegraphics[trim = 87.5mm 0mm 0.5mm 0mm, clip, width=0.16\columnwidth]{figures/reconstructions/wgangp-indep/Arp3_batch0_image0.png}
            &
            \includegraphics[trim = 29.5mm 0mm 58.5mm 0mm, clip, width=0.16\columnwidth]{figures/reconstructions/wgangp-indep/Arp3_batch0_image0.png}
            &
            \includegraphics[trim = 29.5mm 0mm 58.5mm 0mm, clip, width=0.16\columnwidth]{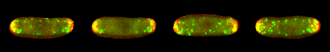}
            &
            \includegraphics[trim = 29.5mm 0mm 58.5mm 0mm, clip, width=0.16\columnwidth]{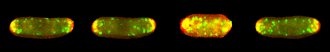}
            &
            \includegraphics[trim = 29.5mm 0mm 58.5mm 0mm, clip, width=0.16\columnwidth]{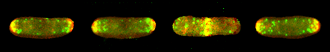} \\[-0.5mm]
            \includegraphics[trim = 0mm 0mm 88mm 0mm, clip, width=0.16\columnwidth]{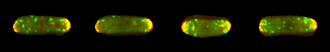}
            &
            \includegraphics[trim = 87.5mm 0mm 0.5mm 0mm, clip, width=0.16\columnwidth]{figures/reconstructions/wgangp-indep/Arp3_batch0_image1.png}
            &
            \includegraphics[trim = 29.5mm 0mm 58.5mm 0mm, clip, width=0.16\columnwidth]{figures/reconstructions/wgangp-indep/Arp3_batch0_image1.png}
            &
            \includegraphics[trim = 29.5mm 0mm 58.5mm 0mm, clip, width=0.16\columnwidth]{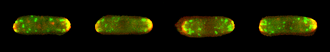}
            &
            \includegraphics[trim = 29.5mm 0mm 58.5mm 0mm, clip, width=0.16\columnwidth]{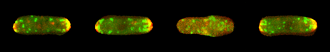}
            &
            \includegraphics[trim = 29.5mm 0mm 58.5mm 0mm, clip, width=0.16\columnwidth]{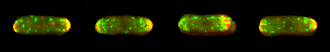} \\[-0.5mm]
            \includegraphics[trim = 0mm 0mm 88mm 0mm, clip, width=0.16\columnwidth]{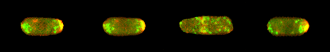}
            &
            \includegraphics[trim = 87.5mm 0mm 0.5mm 0mm, clip, width=0.16\columnwidth]{figures/reconstructions/wgangp-indep/Arp3_batch0_image10.png}
            &
            \includegraphics[trim = 29.5mm 0mm 58.5mm 0mm, clip, width=0.16\columnwidth]{figures/reconstructions/wgangp-indep/Arp3_batch0_image10.png}
            &
            \includegraphics[trim = 29.5mm 0mm 58.5mm 0mm, clip, width=0.16\columnwidth]{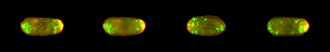}
            &
            \includegraphics[trim = 29.5mm 0mm 58.5mm 0mm, clip, width=0.16\columnwidth]{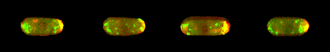}
            &
            \includegraphics[trim = 29.5mm 0mm 58.5mm 0mm, clip, width=0.16\columnwidth]{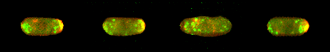} 
            \\[-0.5mm]
            \includegraphics[trim = 0mm 0mm 88mm 0mm, clip, width=0.16\columnwidth]{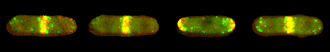}
            &
            \includegraphics[trim = 87.5mm 0mm 0.5mm 0mm, clip, width=0.16\columnwidth]{figures/reconstructions/wgangp-indep/Arp3_batch0_image11.png}
            &
            \includegraphics[trim = 29.5mm 0mm 58.5mm 0mm, clip, width=0.16\columnwidth]{figures/reconstructions/wgangp-indep/Arp3_batch0_image11.png}
            &
            \includegraphics[trim = 29.5mm 0mm 58.5mm 0mm, clip, width=0.16\columnwidth]{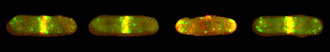}
            &
            \includegraphics[trim = 29.5mm 0mm 58.5mm 0mm, clip, width=0.16\columnwidth]{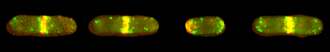}
            &
            \includegraphics[trim = 29.5mm 0mm 58.5mm 0mm, clip, width=0.16\columnwidth]{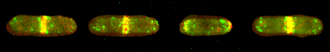} 
            \\[-0.5mm]
            \includegraphics[trim = 0mm 0mm 88mm 0mm, clip, width=0.16\columnwidth]{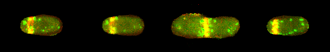}
            &
            \includegraphics[trim = 87.5mm 0mm 0.5mm 0mm, clip, width=0.16\columnwidth]{figures/reconstructions/wgangp-indep/Arp3_batch0_image12.png}
            &
            \includegraphics[trim = 29.5mm 0mm 58.5mm 0mm, clip, width=0.16\columnwidth]{figures/reconstructions/wgangp-indep/Arp3_batch0_image12.png}
            &
            \includegraphics[trim = 29.5mm 0mm 58.5mm 0mm, clip, width=0.16\columnwidth]{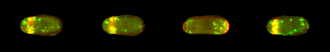}
            &
            \includegraphics[trim = 29.5mm 0mm 58.5mm 0mm, clip, width=0.16\columnwidth]{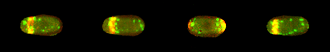}
            &
            \includegraphics[trim = 29.5mm 0mm 58.5mm 0mm, clip, width=0.16\columnwidth]{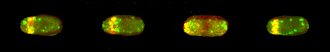}
            \\[-0.5mm]
            \includegraphics[trim = 0mm 0mm 88mm 0mm, clip, width=0.16\columnwidth]{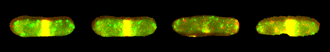}
            &
            \includegraphics[trim = 87.5mm 0mm 0.5mm 0mm, clip, width=0.16\columnwidth]{figures/reconstructions/wgangp-indep/Arp3_batch0_image13.png}
            &
            \includegraphics[trim = 29.5mm 0mm 58.5mm 0mm, clip, width=0.16\columnwidth]{figures/reconstructions/wgangp-indep/Arp3_batch0_image13.png}
            &
            \includegraphics[trim = 29.5mm 0mm 58.5mm 0mm, clip, width=0.16\columnwidth]{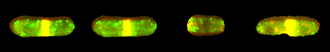}
            &
            \includegraphics[trim = 29.5mm 0mm 58.5mm 0mm, clip, width=0.16\columnwidth]{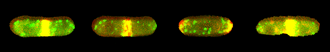}
            &
            \includegraphics[trim = 29.5mm 0mm 58.5mm 0mm, clip, width=0.16\columnwidth]{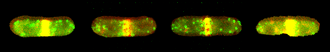}
            \\[-0.5mm]
            \includegraphics[trim = 0mm 0mm 88mm 0mm, clip, width=0.16\columnwidth]{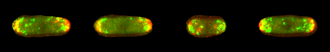}
            &
            \includegraphics[trim = 87.5mm 0mm 0.5mm 0mm, clip, width=0.16\columnwidth]{figures/reconstructions/wgangp-indep/Arp3_batch0_image14.png}
            &
            \includegraphics[trim = 29.5mm 0mm 58.5mm 0mm, clip, width=0.16\columnwidth]{figures/reconstructions/wgangp-indep/Arp3_batch0_image14.png}
            &
            \includegraphics[trim = 29.5mm 0mm 58.5mm 0mm, clip, width=0.16\columnwidth]{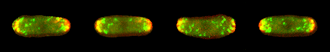}
            &
            \includegraphics[trim = 29.5mm 0mm 58.5mm 0mm, clip, width=0.16\columnwidth]{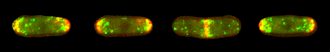}
            &
            \includegraphics[trim = 29.5mm 0mm 58.5mm 0mm, clip, width=0.16\columnwidth]{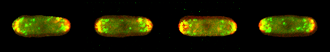}
            \\[2mm]
            \includegraphics[trim = 0mm 0mm 88mm 0mm, clip, width=0.16\columnwidth]{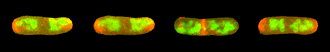}
            &
            \includegraphics[trim = 87.5mm 0mm 0.5mm 0mm, clip, width=0.16\columnwidth]{figures/reconstructions/wgangp-indep/Cki2_batch0_image0.png}
            &
            \includegraphics[trim = 29.5mm 0mm 58.5mm 0mm, clip, width=0.16\columnwidth]{figures/reconstructions/wgangp-indep/Cki2_batch0_image0.png}
            &
            \includegraphics[trim = 29.5mm 0mm 58.5mm 0mm, clip, width=0.16\columnwidth]{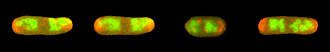}
            &
            \includegraphics[trim = 29.5mm 0mm 58.5mm 0mm, clip, width=0.16\columnwidth]{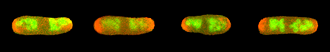}
            &
            \includegraphics[trim = 29.5mm 0mm 58.5mm 0mm, clip, width=0.16\columnwidth]{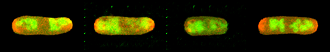} \\[-0.5mm]
            \includegraphics[trim = 0mm 0mm 88mm 0mm, clip, width=0.16\columnwidth]{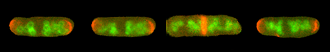}
            &
            \includegraphics[trim = 87.5mm 0mm 0.5mm 0mm, clip, width=0.16\columnwidth]{figures/reconstructions/wgangp-indep/Cki2_batch0_image1.png}
            &
            \includegraphics[trim = 29.5mm 0mm 58.5mm 0mm, clip, width=0.16\columnwidth]{figures/reconstructions/wgangp-indep/Cki2_batch0_image1.png}
            &
            \includegraphics[trim = 29.5mm 0mm 58.5mm 0mm, clip, width=0.16\columnwidth]{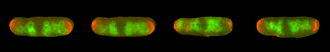}
            &
            \includegraphics[trim = 29.5mm 0mm 58.5mm 0mm, clip, width=0.16\columnwidth]{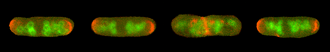}
            &
            \includegraphics[trim = 29.5mm 0mm 58.5mm 0mm, clip, width=0.16\columnwidth]{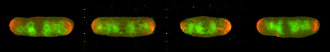} \\[-0.5mm]
            \includegraphics[trim = 0mm 0mm 88mm 0mm, clip, width=0.16\columnwidth]{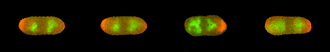}
            &
            \includegraphics[trim = 87.5mm 0mm 0.5mm 0mm, clip, width=0.16\columnwidth]{figures/reconstructions/wgangp-indep/Cki2_batch0_image10.png}
            &
            \includegraphics[trim = 29.5mm 0mm 58.5mm 0mm, clip, width=0.16\columnwidth]{figures/reconstructions/wgangp-indep/Cki2_batch0_image10.png}
            &
            \includegraphics[trim = 29.5mm 0mm 58.5mm 0mm, clip, width=0.16\columnwidth]{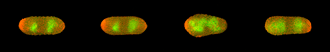}
            &
            \includegraphics[trim = 29.5mm 0mm 58.5mm 0mm, clip, width=0.16\columnwidth]{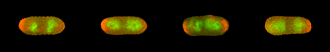}
            &
            \includegraphics[trim = 29.5mm 0mm 58.5mm 0mm, clip, width=0.16\columnwidth]{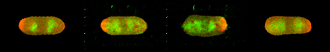} 
            \\[-0.5mm]
            \includegraphics[trim = 0mm 0mm 88mm 0mm, clip, width=0.16\columnwidth]{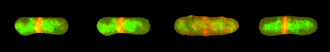}
            &
            \includegraphics[trim = 87.5mm 0mm 0.5mm 0mm, clip, width=0.16\columnwidth]{figures/reconstructions/wgangp-indep/Cki2_batch0_image11.png}
            &
            \includegraphics[trim = 29.5mm 0mm 58.5mm 0mm, clip, width=0.16\columnwidth]{figures/reconstructions/wgangp-indep/Cki2_batch0_image11.png}
            &
            \includegraphics[trim = 29.5mm 0mm 58.5mm 0mm, clip, width=0.16\columnwidth]{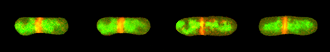}
            &
            \includegraphics[trim = 29.5mm 0mm 58.5mm 0mm, clip, width=0.16\columnwidth]{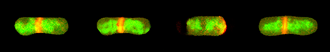}
            &
            \includegraphics[trim = 29.5mm 0mm 58.5mm 0mm, clip, width=0.16\columnwidth]{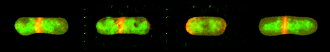} 
            \\[-0.5mm]
            \includegraphics[trim = 0mm 0mm 88mm 0mm, clip, width=0.16\columnwidth]{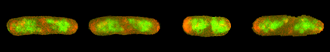}
            &
            \includegraphics[trim = 87.5mm 0mm 0.5mm 0mm, clip, width=0.16\columnwidth]{figures/reconstructions/wgangp-indep/Cki2_batch0_image12.png}
            &
            \includegraphics[trim = 29.5mm 0mm 58.5mm 0mm, clip, width=0.16\columnwidth]{figures/reconstructions/wgangp-indep/Cki2_batch0_image12.png}
            &
            \includegraphics[trim = 29.5mm 0mm 58.5mm 0mm, clip, width=0.16\columnwidth]{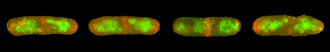}
            &
            \includegraphics[trim = 29.5mm 0mm 58.5mm 0mm, clip, width=0.16\columnwidth]{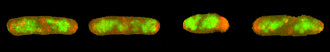}
            &
            \includegraphics[trim = 29.5mm 0mm 58.5mm 0mm, clip, width=0.16\columnwidth]{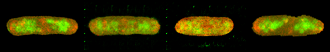} 
            \\[-0.5mm]
            \includegraphics[trim = 0mm 0mm 88mm 0mm, clip, width=0.16\columnwidth]{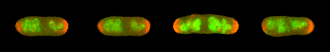}
            &
            \includegraphics[trim = 87.5mm 0mm 0.5mm 0mm, clip, width=0.16\columnwidth]{figures/reconstructions/wgangp-indep/Cki2_batch0_image13.png}
            &
            \includegraphics[trim = 29.5mm 0mm 58.5mm 0mm, clip, width=0.16\columnwidth]{figures/reconstructions/wgangp-indep/Cki2_batch0_image13.png}
            &
            \includegraphics[trim = 29.5mm 0mm 58.5mm 0mm, clip, width=0.16\columnwidth]{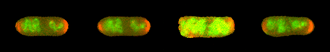}
            &
            \includegraphics[trim = 29.5mm 0mm 58.5mm 0mm, clip, width=0.16\columnwidth]{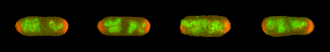}
            &
            \includegraphics[trim = 29.5mm 0mm 58.5mm 0mm, clip, width=0.16\columnwidth]{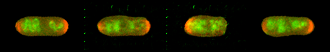}
            \\[-0.5mm]
            \includegraphics[trim = 0mm 0mm 88mm 0mm, clip, width=0.16\columnwidth]{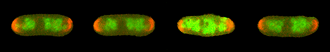}
            &
            \includegraphics[trim = 87.5mm 0mm 0.5mm 0mm, clip, width=0.16\columnwidth]{figures/reconstructions/wgangp-indep/Cki2_batch0_image14.png}
            &
            \includegraphics[trim = 29.5mm 0mm 58.5mm 0mm, clip, width=0.16\columnwidth]{figures/reconstructions/wgangp-indep/Cki2_batch0_image14.png}
            &
            \includegraphics[trim = 29.5mm 0mm 58.5mm 0mm, clip, width=0.16\columnwidth]{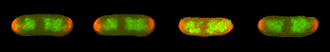}
            &
            \includegraphics[trim = 29.5mm 0mm 58.5mm 0mm, clip, width=0.16\columnwidth]{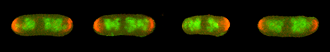}
            &
            \includegraphics[trim = 29.5mm 0mm 58.5mm 0mm, clip, width=0.16\columnwidth]{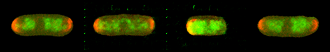} 
            \\[-1mm]
            {\small (a)} &  {\small (b)} & {\small (c)} & {\small (d)} & {\small (e)} & {\small (f)} \\[-2mm]
        \end{tabular}
        &
        \begin{tabular}{@{}c@{\:}c@{\:}c@{\:}c@{\:}c@{\:}c@{}}
            \includegraphics[trim = 0mm 0mm 88mm 0mm, clip, width=0.16\columnwidth]{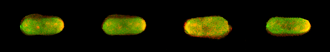}
            &
            \includegraphics[trim = 87.5mm 0mm 0.5mm 0mm, clip, width=0.16\columnwidth]{figures/reconstructions/wgangp-indep/Mkh1_batch0_image0.png}
            &
            \includegraphics[trim = 29.5mm 0mm 58.5mm 0mm, clip, width=0.16\columnwidth]{figures/reconstructions/wgangp-indep/Mkh1_batch0_image0.png}
            &
            \includegraphics[trim = 29.5mm 0mm 58.5mm 0mm, clip, width=0.16\columnwidth]{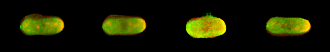}
            &
            \includegraphics[trim = 29.5mm 0mm 58.5mm 0mm, clip, width=0.16\columnwidth]{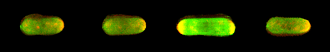}
            &
            \includegraphics[trim = 29.5mm 0mm 58.5mm 0mm, clip, width=0.16\columnwidth]{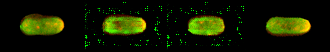} \\[-0.5mm]
            \includegraphics[trim = 0mm 0mm 88mm 0mm, clip, width=0.16\columnwidth]{figures/reconstructions/wgangp-indep/Mkh1_batch0_image1.png}
            &
            \includegraphics[trim = 87.5mm 0mm 0.5mm 0mm, clip, width=0.16\columnwidth]{figures/reconstructions/wgangp-indep/Mkh1_batch0_image1.png}
            &
            \includegraphics[trim = 29.5mm 0mm 58.5mm 0mm, clip, width=0.16\columnwidth]{figures/reconstructions/wgangp-indep/Mkh1_batch0_image1.png}
            &
            \includegraphics[trim = 29.5mm 0mm 58.5mm 0mm, clip, width=0.16\columnwidth]{figures/reconstructions/wgangp-star/Mkh1_batch0_image1.png}
            &
            \includegraphics[trim = 29.5mm 0mm 58.5mm 0mm, clip, width=0.16\columnwidth]{figures/reconstructions/wgangp-star-sep/Mkh1_batch0_image1.png}
            &
            \includegraphics[trim = 29.5mm 0mm 58.5mm 0mm, clip, width=0.16\columnwidth]{figures/reconstructions/gan-star-sep/Mkh1_batch0_image1.png} \\[-0.5mm]
            \includegraphics[trim = 0mm 0mm 88mm 0mm, clip, width=0.16\columnwidth]{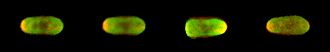}
            &
            \includegraphics[trim = 87.5mm 0mm 0.5mm 0mm, clip, width=0.16\columnwidth]{figures/reconstructions/wgangp-indep/Mkh1_batch0_image10.png}
            &
            \includegraphics[trim = 29.5mm 0mm 58.5mm 0mm, clip, width=0.16\columnwidth]{figures/reconstructions/wgangp-indep/Mkh1_batch0_image10.png}
            &
            \includegraphics[trim = 29.5mm 0mm 58.5mm 0mm, clip, width=0.16\columnwidth]{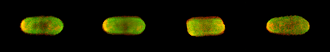}
            &
            \includegraphics[trim = 29.5mm 0mm 58.5mm 0mm, clip, width=0.16\columnwidth]{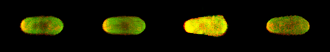}
            &
            \includegraphics[trim = 29.5mm 0mm 58.5mm 0mm, clip, width=0.16\columnwidth]{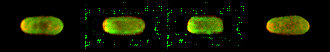} 
            \\[-0.5mm]
            \includegraphics[trim = 0mm 0mm 88mm 0mm, clip, width=0.16\columnwidth]{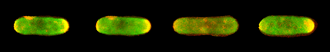}
            &
            \includegraphics[trim = 87.5mm 0mm 0.5mm 0mm, clip, width=0.16\columnwidth]{figures/reconstructions/wgangp-indep/Mkh1_batch0_image11.png}
            &
            \includegraphics[trim = 29.5mm 0mm 58.5mm 0mm, clip, width=0.16\columnwidth]{figures/reconstructions/wgangp-indep/Mkh1_batch0_image11.png}
            &
            \includegraphics[trim = 29.5mm 0mm 58.5mm 0mm, clip, width=0.16\columnwidth]{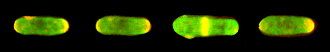}
            &
            \includegraphics[trim = 29.5mm 0mm 58.5mm 0mm, clip, width=0.16\columnwidth]{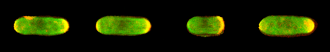}
            &
            \includegraphics[trim = 29.5mm 0mm 58.5mm 0mm, clip, width=0.16\columnwidth]{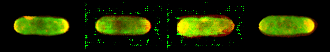} 
            \\[-0.5mm]
            \includegraphics[trim = 0mm 0mm 88mm 0mm, clip, width=0.16\columnwidth]{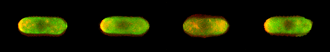}
            &
            \includegraphics[trim = 87.5mm 0mm 0.5mm 0mm, clip, width=0.16\columnwidth]{figures/reconstructions/wgangp-indep/Mkh1_batch0_image12.png}
            &
            \includegraphics[trim = 29.5mm 0mm 58.5mm 0mm, clip, width=0.16\columnwidth]{figures/reconstructions/wgangp-indep/Mkh1_batch0_image12.png}
            &
            \includegraphics[trim = 29.5mm 0mm 58.5mm 0mm, clip, width=0.16\columnwidth]{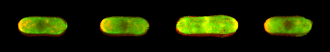}
            &
            \includegraphics[trim = 29.5mm 0mm 58.5mm 0mm, clip, width=0.16\columnwidth]{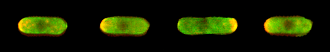}
            &
            \includegraphics[trim = 29.5mm 0mm 58.5mm 0mm, clip, width=0.16\columnwidth]{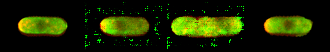}
            \\[-0.5mm]
            \includegraphics[trim = 0mm 0mm 88mm 0mm, clip, width=0.16\columnwidth]{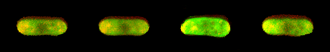}
            &
            \includegraphics[trim = 87.5mm 0mm 0.5mm 0mm, clip, width=0.16\columnwidth]{figures/reconstructions/wgangp-indep/Mkh1_batch0_image13.png}
            &
            \includegraphics[trim = 29.5mm 0mm 58.5mm 0mm, clip, width=0.16\columnwidth]{figures/reconstructions/wgangp-indep/Mkh1_batch0_image13.png}
            &
            \includegraphics[trim = 29.5mm 0mm 58.5mm 0mm, clip, width=0.16\columnwidth]{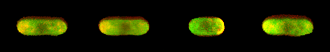}
            &
            \includegraphics[trim = 29.5mm 0mm 58.5mm 0mm, clip, width=0.16\columnwidth]{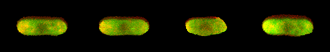}
            &
            \includegraphics[trim = 29.5mm 0mm 58.5mm 0mm, clip, width=0.16\columnwidth]{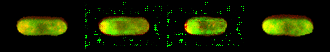}
            \\[-0.5mm]
            \includegraphics[trim = 0mm 0mm 88mm 0mm, clip, width=0.16\columnwidth]{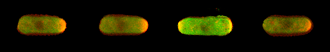}
            &
            \includegraphics[trim = 87.5mm 0mm 0.5mm 0mm, clip, width=0.16\columnwidth]{figures/reconstructions/wgangp-indep/Mkh1_batch0_image14.png}
            &
            \includegraphics[trim = 29.5mm 0mm 58.5mm 0mm, clip, width=0.16\columnwidth]{figures/reconstructions/wgangp-indep/Mkh1_batch0_image14.png}
            &
            \includegraphics[trim = 29.5mm 0mm 58.5mm 0mm, clip, width=0.16\columnwidth]{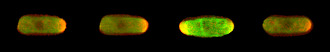}
            &
            \includegraphics[trim = 29.5mm 0mm 58.5mm 0mm, clip, width=0.16\columnwidth]{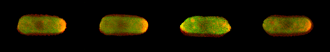}
            &
            \includegraphics[trim = 29.5mm 0mm 58.5mm 0mm, clip, width=0.16\columnwidth]{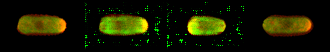} 
            \\[2mm]
            \includegraphics[trim = 0mm 0mm 88mm 0mm, clip, width=0.16\columnwidth]{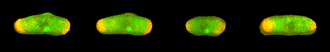}
            &
            \includegraphics[trim = 87.5mm 0mm 0.5mm 0mm, clip, width=0.16\columnwidth]{figures/reconstructions/wgangp-indep/Sid2_batch0_image0.png}
            &
            \includegraphics[trim = 29.5mm 0mm 58.5mm 0mm, clip, width=0.16\columnwidth]{figures/reconstructions/wgangp-indep/Sid2_batch0_image0.png}
            &
            \includegraphics[trim = 29.5mm 0mm 58.5mm 0mm, clip, width=0.16\columnwidth]{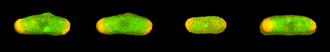}
            &
            \includegraphics[trim = 29.5mm 0mm 58.5mm 0mm, clip, width=0.16\columnwidth]{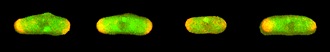}
            &
            \includegraphics[trim = 29.5mm 0mm 58.5mm 0mm, clip, width=0.16\columnwidth]{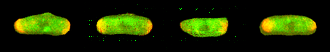} \\[-0.5mm]
            \includegraphics[trim = 0mm 0mm 88mm 0mm, clip, width=0.16\columnwidth]{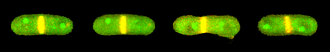}
            &
            \includegraphics[trim = 87.5mm 0mm 0.5mm 0mm, clip, width=0.16\columnwidth]{figures/reconstructions/wgangp-indep/Sid2_batch0_image1.png}
            &
            \includegraphics[trim = 29.5mm 0mm 58.5mm 0mm, clip, width=0.16\columnwidth]{figures/reconstructions/wgangp-indep/Sid2_batch0_image1.png}
            &
            \includegraphics[trim = 29.5mm 0mm 58.5mm 0mm, clip, width=0.16\columnwidth]{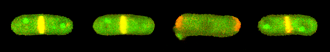}
            &
            \includegraphics[trim = 29.5mm 0mm 58.5mm 0mm, clip, width=0.16\columnwidth]{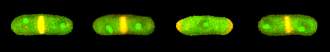}
            &
            \includegraphics[trim = 29.5mm 0mm 58.5mm 0mm, clip, width=0.16\columnwidth]{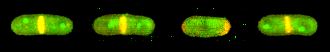} \\[-0.5mm]
            \includegraphics[trim = 0mm 0mm 88mm 0mm, clip, width=0.16\columnwidth]{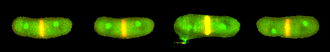}
            &
            \includegraphics[trim = 87.5mm 0mm 0.5mm 0mm, clip, width=0.16\columnwidth]{figures/reconstructions/wgangp-indep/Sid2_batch0_image10.png}
            &
            \includegraphics[trim = 29.5mm 0mm 58.5mm 0mm, clip, width=0.16\columnwidth]{figures/reconstructions/wgangp-indep/Sid2_batch0_image10.png}
            &
            \includegraphics[trim = 29.5mm 0mm 58.5mm 0mm, clip, width=0.16\columnwidth]{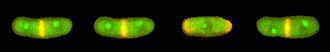}
            &
            \includegraphics[trim = 29.5mm 0mm 58.5mm 0mm, clip, width=0.16\columnwidth]{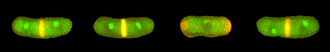}
            &
            \includegraphics[trim = 29.5mm 0mm 58.5mm 0mm, clip, width=0.16\columnwidth]{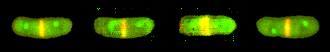} 
            \\[-0.5mm]
            \includegraphics[trim = 0mm 0mm 88mm 0mm, clip, width=0.16\columnwidth]{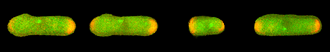}
            &
            \includegraphics[trim = 87.5mm 0mm 0.5mm 0mm, clip, width=0.16\columnwidth]{figures/reconstructions/wgangp-indep/Sid2_batch0_image11.png}
            &
            \includegraphics[trim = 29.5mm 0mm 58.5mm 0mm, clip, width=0.16\columnwidth]{figures/reconstructions/wgangp-indep/Sid2_batch0_image11.png}
            &
            \includegraphics[trim = 29.5mm 0mm 58.5mm 0mm, clip, width=0.16\columnwidth]{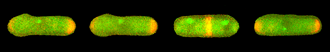}
            &
            \includegraphics[trim = 29.5mm 0mm 58.5mm 0mm, clip, width=0.16\columnwidth]{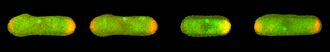}
            &
            \includegraphics[trim = 29.5mm 0mm 58.5mm 0mm, clip, width=0.16\columnwidth]{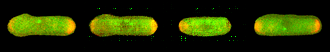} 
            \\[-0.5mm]
            \includegraphics[trim = 0mm 0mm 88mm 0mm, clip, width=0.16\columnwidth]{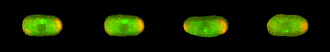}
            &
            \includegraphics[trim = 87.5mm 0mm 0.5mm 0mm, clip, width=0.16\columnwidth]{figures/reconstructions/wgangp-indep/Sid2_batch0_image12.png}
            &
            \includegraphics[trim = 29.5mm 0mm 58.5mm 0mm, clip, width=0.16\columnwidth]{figures/reconstructions/wgangp-indep/Sid2_batch0_image12.png}
            &
            \includegraphics[trim = 29.5mm 0mm 58.5mm 0mm, clip, width=0.16\columnwidth]{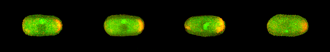}
            &
            \includegraphics[trim = 29.5mm 0mm 58.5mm 0mm, clip, width=0.16\columnwidth]{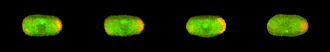}
            &
            \includegraphics[trim = 29.5mm 0mm 58.5mm 0mm, clip, width=0.16\columnwidth]{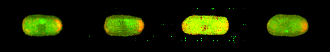}
            \\[-0.5mm]
            \includegraphics[trim = 0mm 0mm 88mm 0mm, clip, width=0.16\columnwidth]{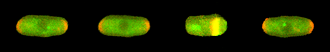}
            &
            \includegraphics[trim = 87.5mm 0mm 0.5mm 0mm, clip, width=0.16\columnwidth]{figures/reconstructions/wgangp-indep/Sid2_batch0_image13.png}
            &
            \includegraphics[trim = 29.5mm 0mm 58.5mm 0mm, clip, width=0.16\columnwidth]{figures/reconstructions/wgangp-indep/Sid2_batch0_image13.png}
            &
            \includegraphics[trim = 29.5mm 0mm 58.5mm 0mm, clip, width=0.16\columnwidth]{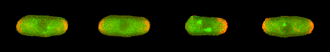}
            &
            \includegraphics[trim = 29.5mm 0mm 58.5mm 0mm, clip, width=0.16\columnwidth]{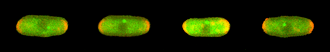}
            &
            \includegraphics[trim = 29.5mm 0mm 58.5mm 0mm, clip, width=0.16\columnwidth]{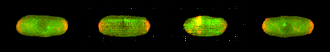}
            \\[-0.5mm]
            \includegraphics[trim = 0mm 0mm 88mm 0mm, clip, width=0.16\columnwidth]{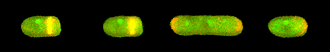}
            &
            \includegraphics[trim = 87.5mm 0mm 0.5mm 0mm, clip, width=0.16\columnwidth]{figures/reconstructions/wgangp-indep/Sid2_batch0_image14.png}
            &
            \includegraphics[trim = 29.5mm 0mm 58.5mm 0mm, clip, width=0.16\columnwidth]{figures/reconstructions/wgangp-indep/Sid2_batch0_image14.png}
            &
            \includegraphics[trim = 29.5mm 0mm 58.5mm 0mm, clip, width=0.16\columnwidth]{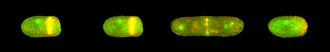}
            &
            \includegraphics[trim = 29.5mm 0mm 58.5mm 0mm, clip, width=0.16\columnwidth]{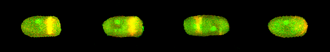}
            &
            \includegraphics[trim = 29.5mm 0mm 58.5mm 0mm, clip, width=0.16\columnwidth]{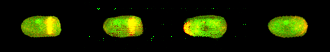}
            \\[2mm]
            \includegraphics[trim = 0mm 0mm 88mm 0mm, clip, width=0.16\columnwidth]{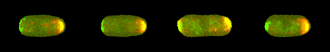}
            &
            \includegraphics[trim = 87.5mm 0mm 0.5mm 0mm, clip, width=0.16\columnwidth]{figures/reconstructions/wgangp-indep/Tea1_batch0_image0.png}
            &
            \includegraphics[trim = 29.5mm 0mm 58.5mm 0mm, clip, width=0.16\columnwidth]{figures/reconstructions/wgangp-indep/Tea1_batch0_image0.png}
            &
            \includegraphics[trim = 29.5mm 0mm 58.5mm 0mm, clip, width=0.16\columnwidth]{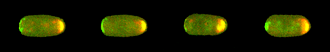}
            &
            \includegraphics[trim = 29.5mm 0mm 58.5mm 0mm, clip, width=0.16\columnwidth]{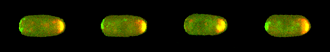}
            &
            \includegraphics[trim = 29.5mm 0mm 58.5mm 0mm, clip, width=0.16\columnwidth]{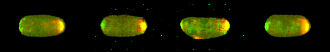} \\[-0.5mm]
            \includegraphics[trim = 0mm 0mm 88mm 0mm, clip, width=0.16\columnwidth]{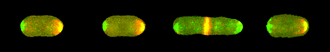}
            &
            \includegraphics[trim = 87.5mm 0mm 0.5mm 0mm, clip, width=0.16\columnwidth]{figures/reconstructions/wgangp-indep/Tea1_batch0_image1.png}
            &
            \includegraphics[trim = 29.5mm 0mm 58.5mm 0mm, clip, width=0.16\columnwidth]{figures/reconstructions/wgangp-indep/Tea1_batch0_image1.png}
            &
            \includegraphics[trim = 29.5mm 0mm 58.5mm 0mm, clip, width=0.16\columnwidth]{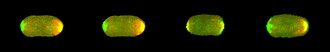}
            &
            \includegraphics[trim = 29.5mm 0mm 58.5mm 0mm, clip, width=0.16\columnwidth]{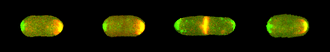}
            &
            \includegraphics[trim = 29.5mm 0mm 58.5mm 0mm, clip, width=0.16\columnwidth]{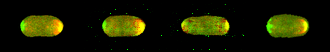} \\[-0.5mm]
            \includegraphics[trim = 0mm 0mm 88mm 0mm, clip, width=0.16\columnwidth]{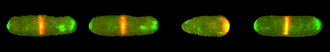}
            &
            \includegraphics[trim = 87.5mm 0mm 0.5mm 0mm, clip, width=0.16\columnwidth]{figures/reconstructions/wgangp-indep/Tea1_batch0_image10.png}
            &
            \includegraphics[trim = 29.5mm 0mm 58.5mm 0mm, clip, width=0.16\columnwidth]{figures/reconstructions/wgangp-indep/Tea1_batch0_image10.png}
            &
            \includegraphics[trim = 29.5mm 0mm 58.5mm 0mm, clip, width=0.16\columnwidth]{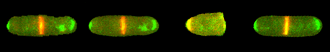}
            &
            \includegraphics[trim = 29.5mm 0mm 58.5mm 0mm, clip, width=0.16\columnwidth]{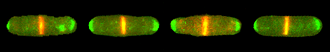}
            &
            \includegraphics[trim = 29.5mm 0mm 58.5mm 0mm, clip, width=0.16\columnwidth]{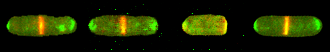} 
            \\[-0.5mm]
            \includegraphics[trim = 0mm 0mm 88mm 0mm, clip, width=0.16\columnwidth]{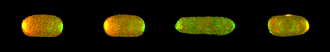}
            &
            \includegraphics[trim = 87.5mm 0mm 0.5mm 0mm, clip, width=0.16\columnwidth]{figures/reconstructions/wgangp-indep/Tea1_batch0_image11.png}
            &
            \includegraphics[trim = 29.5mm 0mm 58.5mm 0mm, clip, width=0.16\columnwidth]{figures/reconstructions/wgangp-indep/Tea1_batch0_image11.png}
            &
            \includegraphics[trim = 29.5mm 0mm 58.5mm 0mm, clip, width=0.16\columnwidth]{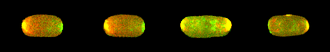}
            &
            \includegraphics[trim = 29.5mm 0mm 58.5mm 0mm, clip, width=0.16\columnwidth]{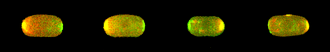}
            &
            \includegraphics[trim = 29.5mm 0mm 58.5mm 0mm, clip, width=0.16\columnwidth]{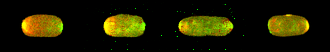} 
            \\[-0.5mm]
            \includegraphics[trim = 0mm 0mm 88mm 0mm, clip, width=0.16\columnwidth]{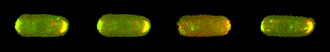}
            &
            \includegraphics[trim = 87.5mm 0mm 0.5mm 0mm, clip, width=0.16\columnwidth]{figures/reconstructions/wgangp-indep/Tea1_batch0_image12.png}
            &
            \includegraphics[trim = 29.5mm 0mm 58.5mm 0mm, clip, width=0.16\columnwidth]{figures/reconstructions/wgangp-indep/Tea1_batch0_image12.png}
            &
            \includegraphics[trim = 29.5mm 0mm 58.5mm 0mm, clip, width=0.16\columnwidth]{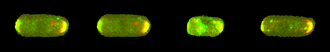}
            &
            \includegraphics[trim = 29.5mm 0mm 58.5mm 0mm, clip, width=0.16\columnwidth]{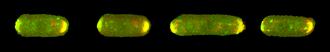}
            &
            \includegraphics[trim = 29.5mm 0mm 58.5mm 0mm, clip, width=0.16\columnwidth]{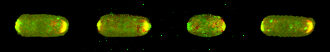}
            \\[-0.5mm]
            \includegraphics[trim = 0mm 0mm 88mm 0mm, clip, width=0.16\columnwidth]{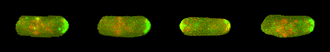}
            &
            \includegraphics[trim = 87.5mm 0mm 0.5mm 0mm, clip, width=0.16\columnwidth]{figures/reconstructions/wgangp-indep/Tea1_batch0_image13.png}
            &
            \includegraphics[trim = 29.5mm 0mm 58.5mm 0mm, clip, width=0.16\columnwidth]{figures/reconstructions/wgangp-indep/Tea1_batch0_image13.png}
            &
            \includegraphics[trim = 29.5mm 0mm 58.5mm 0mm, clip, width=0.16\columnwidth]{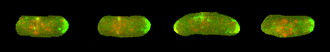}
            &
            \includegraphics[trim = 29.5mm 0mm 58.5mm 0mm, clip, width=0.16\columnwidth]{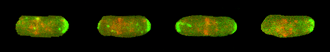}
            &
            \includegraphics[trim = 29.5mm 0mm 58.5mm 0mm, clip, width=0.16\columnwidth]{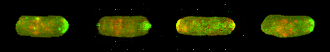}
            \\[-0.5mm]
            \includegraphics[trim = 0mm 0mm 88mm 0mm, clip, width=0.16\columnwidth]{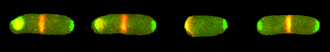}
            &
            \includegraphics[trim = 87.5mm 0mm 0.5mm 0mm, clip, width=0.16\columnwidth]{figures/reconstructions/wgangp-indep/Tea1_batch0_image14.png}
            &
            \includegraphics[trim = 29.5mm 0mm 58.5mm 0mm, clip, width=0.16\columnwidth]{figures/reconstructions/wgangp-indep/Tea1_batch0_image14.png}
            &
            \includegraphics[trim = 29.5mm 0mm 58.5mm 0mm, clip, width=0.16\columnwidth]{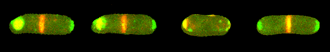}
            &
            \includegraphics[trim = 29.5mm 0mm 58.5mm 0mm, clip, width=0.16\columnwidth]{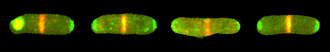}
            &
            \includegraphics[trim = 29.5mm 0mm 58.5mm 0mm, clip, width=0.16\columnwidth]{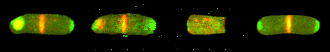} 
            \\[-1mm]
            {\small (a)} &  {\small (b)} & {\small (c)} & {\small (d)} & {\small (e)} & {\small (f)} \\[-2mm]
        \end{tabular}
        \end{tabular}
    \end{center}
    \caption{Examples of cell reconstructions (an extension of Figure \ref{fig:reconstructionExamples}).
        (a)~-- a test image; (b)~-- the $L_2$ nearest neighbor; (c)~-- regular reconstruction by one-class separable WGAN-GP; (d)~-- regular reconstruction by star-shaped WGAN-GP; (e)~-- separable reconstruction by star-shaped WGAN-GP; (f)~-- separable reconstruction by star-shaped GAN.
        \label{fig:reconstructionExamples_all}}
\end{figure*}

Figures~\ref{fig:reconstruction_all} and~\ref{fig:reconstructionExamples_all} show extended results of the reconstruction experiment described in Section~\ref{sec:exp:optimization}. We show scatter plots of NLL vs. reconstruction error for the six selected proteins and provide reconstruction examples for all of them.

\end{document}